\documentclass[10pt]{config/paper}

\title{COLA: Characterizing and Optimizing the Tail Latency\\ for Safe Level-4 Autonomous Vehicle Systems}

\author{Haolan Liu \qquad Zixuan Wang \qquad Jishen Zhao \\ University of California, San Diego}

\newcommand{\hint}[2]{\vspace{2pt}\noindent\textbf{#1}{#2}\vspace{2pt}}

\begin{document}


\maketitle


\begin{abstract}
    Autonomous vehicles (AVs)
    are envisioned to revolutionize our life by providing safe, relaxing, and convenient ground transportation.
    The computing systems in such vehicles are required to interpret various sensor data and generate responses to the environment in a timely manner to ensure driving safety.
    However, such timing-related safety requirements are largely unexplored in prior works.

    In this paper, we conduct a systematic study to understand the timing requirements of AV systems.
    We focus on investigating and mitigating the sources of tail latency 
    in Level-4 AV computing systems. We observe that the performance of AV algorithms is not uniformly distributed -- instead, the latency is susceptible to vehicle environment fluctuations, such as traffic density.
    This contributes to burst computation and memory access in response to the traffic, and further leads to tail latency in the system.
    Furthermore, we observe that tail latency also comes from a mismatch between the pre-configured AV computation pipeline and the dynamic latency requirements in real-world driving scenarios.
    
    Based on these observations, we propose a set of system designs to mitigate AV tail latency. We demonstrate our design on widely-used industrial Level-4 AV systems, Baidu Apollo and Autoware.
    The evaluation shows our design achieves 1.65$\times$ improvement over the worst-case latency and 1.3$\times$ over the average latency, and avoids 93\% of accidents in Apollo.
\end{abstract}


\section{Introduction}


The fast-growing autonomous vehicle (AV) technology is leading toward a paradigm shift in driving safety.
In traditional human-centered traffic, human faults account for over 90\% of
accidents~\cite{crashsurvey}, while bad driving habits are the 
major reason for traffic congestion~\cite{congestionsurvey}.
By offloading driving tasks to computing systems, AVs are envisioned to
significantly reduce the occurrence of car accidents and fatalities while improving traffic efficiency~\cite{nhtsasafety}.
Compared to human drivers, AV systems can achieve fully-attentive driving, 360-degree awareness, and shorter reaction time~\cite{nhtsasafety}.


AV systems are categorized into 6 levels~\cite{TaxonomyAD} based on their driving automation and safety, ranging from level 0 (fully manual) to 5 (fully autonomous).
Level-4 AVs mark the huge breakthrough of requiring no human drivers~\footnote{Level-4 AVs can operate in certain areas, while Level-5 AVs can drive to anywhere.}, while the lower-level AV needs a safety driver to take over during an emergency.
To ensure safety, AV systems need to make timely driving decisions in response to complicated and highly dynamic real-world driving environments.

\haolan{Describe briefly the real-world complexities.}

To cope with the huge complexities of real-world driving tasks,
Level-4 AVs are equipped with various high-profile sensors, such as high-resolution cameras and LiDARs~\cite{9304602}, which generate large volumes of data to be fused and processed by high-performance server-like in-vehicle computing systems;
the computing system typically works as a deep pipeline with many software modules: it proceeds with periodic sensor data and generates driving commands (e.g., steer, accelerate, and break)~\cite{10.1145/3296957.3173191,Luo2019TimeCA,10.1145/3492321.3519576,Gog2021PylotAM}.

However, such pipelined computation is more sensitive to tail latency effects than conventional real-time systems due to latency aggregation.
Furthermore, the characteristic of tail latency and the impact on computation varies in a dynamic manner affected by driving scenarios. 
A safety-critical module may occasionally incur long computation latency,
accumulating with bursts and leading to slow reactions to a dynamically changing traffic environment~\cite{Luo2019TimeCA,gan2020eudoxus}.
Recent studies on AV accidents suggest that such slow reactions are likely to cause AVs to fail to yield to pedestrians or stop at stop signs~\cite{uberaccident,teslaaccident,teslaaccident2,teslaaccident3}.

To better understand the causes of tail latency and identify potential opportunities for improvement,
we propose COLA
, a framework that systematically characterizes and optimizes the tail latency effect in real-world Level-4 AV systems.
COLA comprises a set of driving scenario data and profiling tools for tail latency analysis.
COLA 
can work with
popular full-stack AV driving platforms (such as Baidu Apollo~\cite{apollo} and Autoware~\cite{Bateni2019PredictableDR}) 
Since those platforms select their algorithms based on individual design concerns, they only integrate quite a narrow range of AV algorithms.
To this end, we also integrate individual AV algorithm libraries (such as MMCV~\cite{mmcv} and Detectron~\cite{wu2019detectron2}).
Despite the importance, to our knowledge,
no public studies investigated the
tail latency in Level-4 AV systems in a comprehensive manner.

Using COLA, we perform a detailed tail latency analysis mainly on Baidu Apollo, one of the most widely-used AV platforms~\cite{apollo}.
We make seven key observations in three categories.
\emph{Category 1}: Investigating and modeling the AV reaction time exposes issues with fixed dataflow and entangled priorities that slowdown AV reaction time, introducing safety risks in near collision scenarios.
\emph{Category 2}: Tail latency characterization shows (i) the impact of traffic patterns and heaviness on the variation of latency characteristics and requirements, (ii) the challenge with predictive computation in AV software modules, (iii) and a mismatch between fixed AV algorithm configurations and dynamic latency requirements in different driving scenarios.
\emph{Category 3}: System-level factors, such as serious resource contention, low computation utilization, and suboptimal system throughput imposed by traditional resource allocation and scheduling schemes.

Based on the characterization and observations, we design a set of
system techniques
to mitigate the impact of tail latency on Level-4 AV systems. First, to allow AVs to react fast in dangerous scenarios, we propose an adaptive data flow, which adopts fast paths and a downstream mechanism to make AV systems adaptive in
different driving scenarios. Second, we develop a proactive processing scheme to alleviate the latency variation in AV systems. Third, we propose a best-effort work stealing scheme that allows the operating system and runtime to leverage the bursty pattern in heavy traffic for resource scheduling.
Extensive experimental evaluation shows that our design is highly efficient with an average latency of 52.38 ms and worst-case latency of 123.93 ms, compared with a baseline latency of 73.73 ms and 203.16 ms in the worst case.
We also validate the safety of our design in high-fidelity simulators.
Our results show that COLA avoids 93\% of accidents.

This paper makes the following contributions:
\squishlist
\item We propose COLA, the first framework for comprehensive characterization of tail latency in Level-4 AV systems. We systematically study and collect multiple AV implementations in COLA.
Moreover, COLA also provides a diverse set of driving scenarios and profiling tools for AV designers to understand the sources of tail latency.
\item We extensively profile the state-of-the-art AV platform and algorithms to investigate the characteristics and causes of tail latency and make seven key observations.
\item Based on the observations, we propose a set of AV system design principles
to mitigate the impact of tail latency on system performance and safety.


\squishend

\section{Background}


\subsection{Level-4 AVs}

Future AVs are envisioned to be highly automotive and do not require human input in most circumstances.
Such AVs, defined as Level-4 (or \emph{high automation}) AVs, are under active development by corporations including Google~\cite{waymo}, Uber~\cite{UberL4}, 
and Baidu~\cite{apollodebut}. As a comparison, low-level AVs (Level 1-3) require drivers' operation in certain conditions.

\begin{figure}[t]
    \centering
    \includegraphics[width=\linewidth]{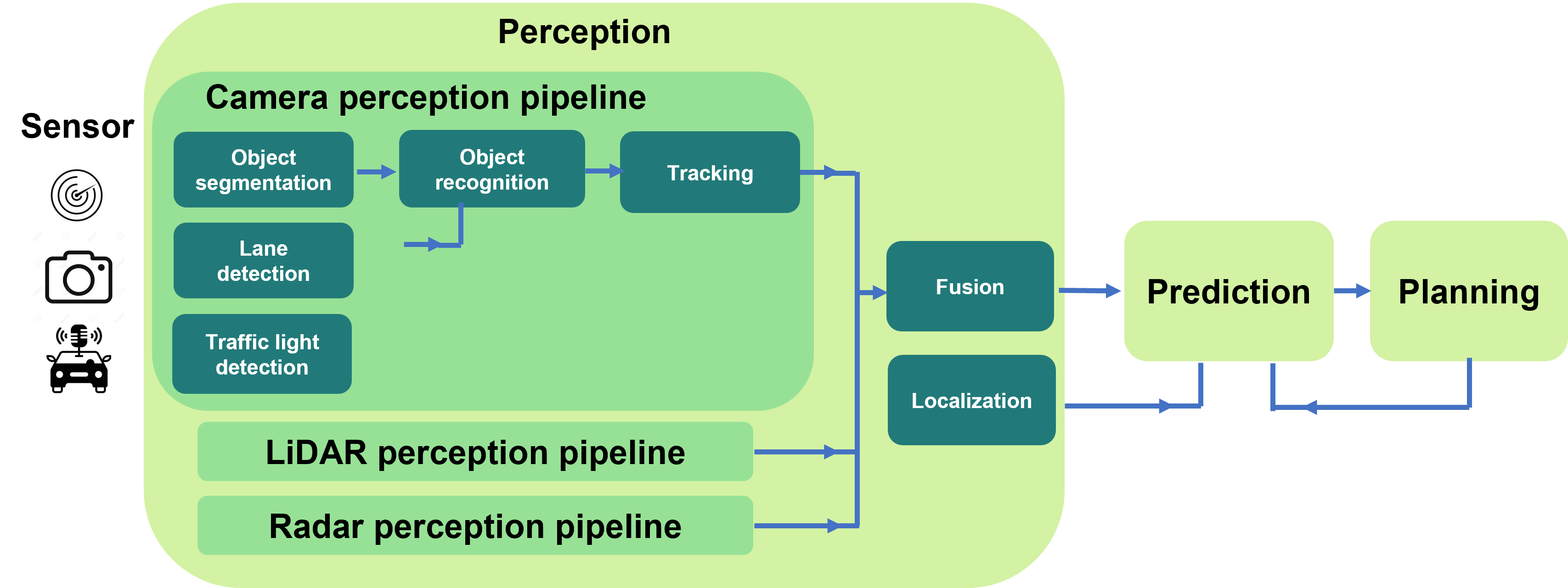}
    \caption{A typical modular pipeline of a Level-4 AV system. We also show the internal structure of the perception module and its complex topology.}
    \label{fig:av_overview}
    \vspace{-15pt}
\end{figure}

Typically, Level-4 AV system decomposes the driving task into perception, prediction, and planning.
Figure~\ref{fig:av_overview} presents a high-level overview of the Baidu Apollo system, where multiple sensors and computation modules are organized as a pipeline, including sensor hardware and computation software modules.
The sensors (e.g., LiDAR, camera, and radar) periodically stream data to the computation pipeline, to update the current status of the vehicle and its environment.
Based on the sensor data, the \emph{perception} module perceives the surrounding environments, e.g., recognizing nearby objects such as cyclists and other vehicles.
The results are fed to the \emph{prediction} module to predict the future action of dynamic objects including vehicles and pedestrians. 
The \emph{planning} module determines the optimal driving decisions in terms of vehicle efficiency, riding experiences, and safety.
And finally, the control actuation module will translate these decisions into vehicle control signals and sends them to the corresponding vehicle hardware.

Figure~\ref{fig:av_overview} also shows the complex internal structure of the high-level module (The figure only shows the perception module).
In practice, an AV computing system comprises hundreds of interdependent submodules~\cite{Luo2019TimeCA,zhao2019towards}, including a diverse range of workloads such as image and point cloud processing, model inference, and numerical optimization. 

As AV systems have increasingly high computation loads and complex data pipelines to accommodate complex driving tasks, vendors commonly adopt high-performance server-like in-vehicle computing systems, including CPUs, GPUs, FPGAs, storage devices, and customized accelerators~\cite{9304602}. Yet the onboard hardware is limited by energy consumption, heat dissipation, costs, and also weights~\cite{254354, 10.1145/3296957.3173191}; It is challenging to utilize limited onboard hardware efficiently while still guaranteeing realtime performance.

\subsection{AV Latency Requirements}
\label{section:latencyrequirements}

Tail latency degrades the AV system safety by generating stale input for downstream modules and delaying the driving decisions.
Despite many attempts being made to define or formalize such latency requirements for AVs~\cite{Luo2019TimeCA, 10.1145/3296957.3173191},  there is no single well-defined standard for all scenarios.
In this section, we present a quantitative analysis of the factors that impact the maximum latency requirements to avoid crashes.

\begin{figure}[t]
    \centering
    \includegraphics[width=0.45\textwidth]{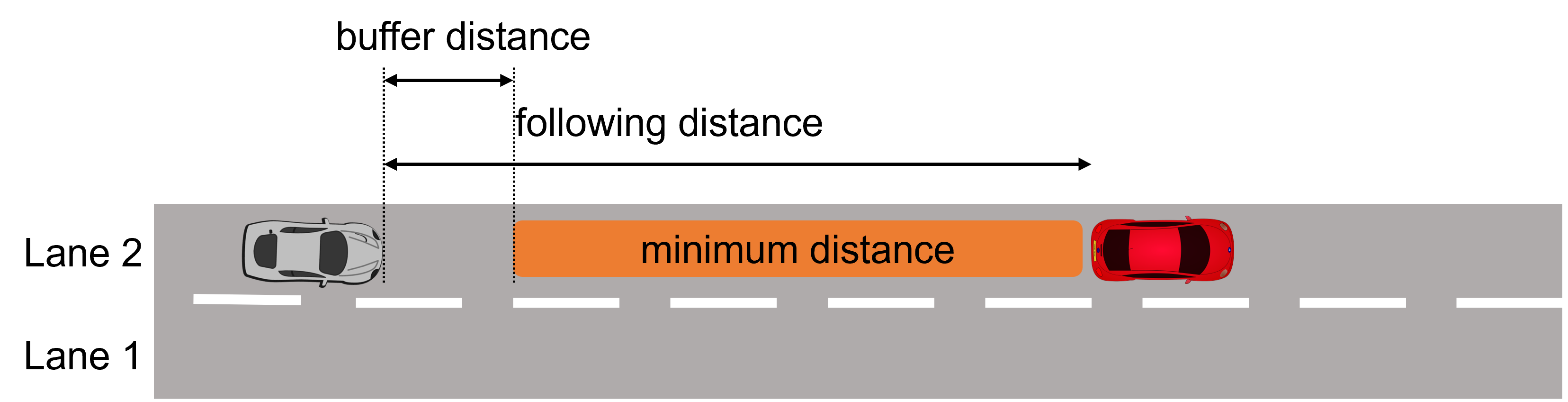}
    \caption{A typical vehicle following scenario.}
    \vspace{-10pt}
    \label{fig:scenario1}
\end{figure}

\para{Dynamic Latency Requirements.} AVs have dynamic latency requirements in terms of distances and velocities. Figure~\ref{fig:scenario1} shows a vehicle-following driving scenario:
the AV (grey vehicle) is driving behind the red vehicle, while the red vehicle suddenly decelerates.
The AV should be able to quickly detect such behavior and make decisions to avoid crashes.
According to the reports in commercial AVs, at a typical speed of 5.6 m/s, an average of 164 ms latency is able to avoid obstacles that are 5 m away, while a latency tail (740 ms) can only avoid obstacles 8.3 m away~\cite{9251973}.

To describe the latency requirements formally, responsibility-sensitive safety (RSS) 
defines a \emph{safety envelope} around the AV~\cite{ShalevShwartz2017OnAF}: the minimum longitudinal and lateral distance that should be maintained to avoid danger. The orange area in Figure~\ref{fig:scenario1} is the minimum longitudinal distance.
In this scenario, when the preceding vehicle decelerates, the AV takes time $t_{reaction}$ to respond (decelerate).
After $t_{reaction}$, the relative distance will decrease by $d(t_{reaction})$. 
As the AV needs to stay out of the minimum distance or it may collide with the preceding vehicle, we can calculate $t_{reaction}$ based on:

\begin{align}
    d \left( t_{reaction} \right) < d_{buffer}
\end{align}

$d_{buffer}$ is the buffer distance in Figure~\ref{fig:scenario1}. $d(t)$ is a function that calculates the distance after time $t$ 
based on physical laws and the current physical states of two vehicles (positions, velocities, and acceleration).
The figure describes the longitudinal (the road direction) dangerous threshold, while the lateral dangerous threshold can also be defined similarly.

\begin{figure}[t]
    \centering
    \includegraphics[width=0.45\textwidth]{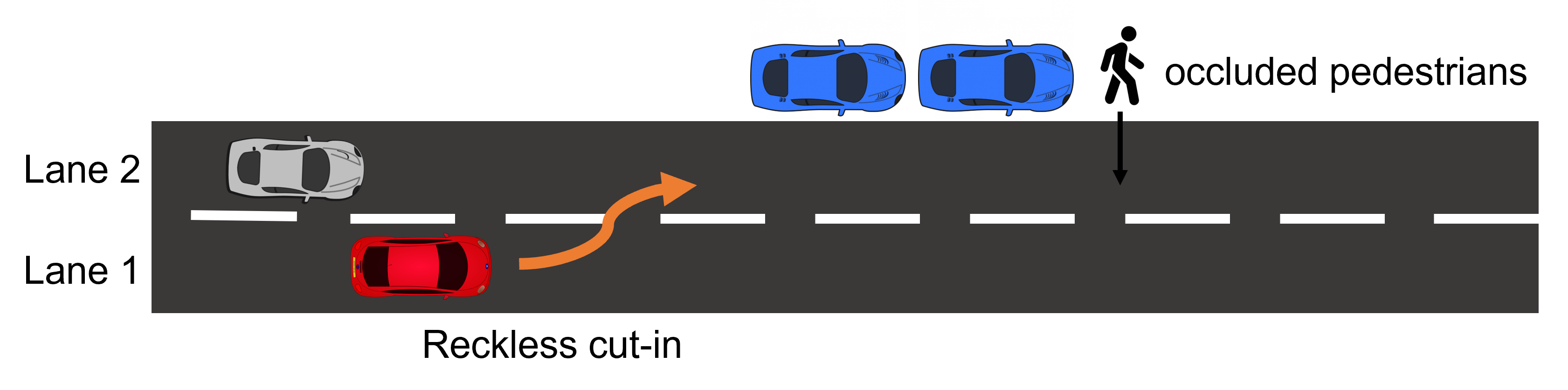}
    \caption{Two corner-case scenarios that require faster reaction time.}
    \vspace{-10pt}
    \label{fig:scenario2}
\end{figure}

\para{Corner-case Driving Scenarios.} 
Driving scenarios represent the environment around the AV, including road conditions, traffic lights, buildings, vehicles, and pedestrians.
Corner-case driving scenarios, although rarely happen, should be included in the latency analysis to maintain safety.
E.g., Figure~\ref{fig:scenario2} shows two such corner cases: aggressive cut-in and occlusion.
The first scenario is that the red vehicle recklessly cuts in from Lane 2 without leaving enough space. Although it is not the responsibility of AVs to avoid crashes~\cite{ShalevShwartz2017OnAF}, the AV should still try to keep safe if possible.
The second scenario is the crossing pedestrian occluded by parked vehicles. As the pedestrian cannot be detected before it shows up, AVs are given a much shorter time window to react.

To quantitatively study such latency requirements, we also analyze four driving scenarios and corresponding latency requirements to avoid crashes, as listed in Table~\ref{table:scenarioqos}. In the vehicle-following scenario, the preceding vehicle suddenly decelerates and the AV needs to stop immediately to avoid crashes.
We delay the control command with increasing latency until observing the crash. We deem the threshold end-to-end latency as the latency requirement for the scenario.
The above analysis shows that the latency requirements of AVs depend on many factors in driving scenarios (including road conditions and other vehicles' behavior).


\begin{table}[t!]
  \centering
    \begin{tabular}{lccc}
        \toprule
        \multirow{2}{*}{Category} & \multicolumn{1}{c}{AV Speed} &\multicolumn{1}{c}{Distance} & \multicolumn{1}{c}{Latency} \\
        ~ & \multicolumn{1}{c}{(km/h)} & \multicolumn{1}{c}{(m)} & \multicolumn{1}{c}{(ms)} \\
        \midrule
        Vehicle Following  & 35.0 & 10.0 & 411.2 \\
        Vehicle Following  & 20.0 & 10.0 & 621.8 \\
        Encroaching Cut-in & 25.0 &  4.7 & 235.5 \\
        Occluded Cut-in    & 25.0 &  3.9 & 159.5 \\
        \bottomrule
    \end{tabular}
    \caption{Estimated latency requirements in different driving scenarios. }
    \label{table:scenarioqos}
    \vspace{-20pt}
\end{table}

\subsection{Bursty Workload}

\begin{figure}[t]
    \centering
    \includegraphics[width=0.40\textwidth]{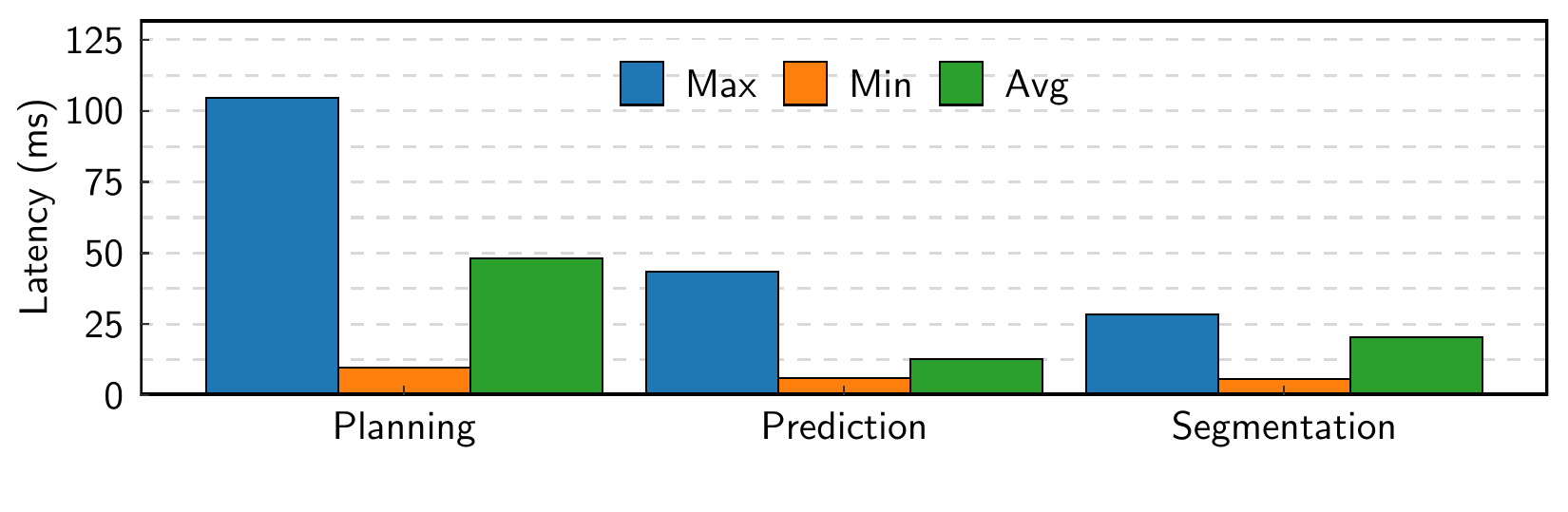}
        \vspace{-15pt}
    \caption{Latency in different Apollo modules.}
        \vspace{-10pt}
    \label{fig:bursty}
\end{figure}

\para{Large Latency Variation.}
AV system exhibits large latency variation.
Figure~\ref{fig:bursty} shows our latency analysis of four AV modules under various driving scenarios.
Our profiling shows that the maximum latency in the planning module is 10 $\times$ of the minimum latency.
When the AV is driving in heavier and more complex traffic, the planning module needs more computation to analyze the environment and handle the uncertainty.
We also observed similar bursty latency effects in the performance of the segmentation module and the planning module.
We provide a more detailed analysis of the burstiness and its root reason in Section~\ref{section:source}.


\begin{figure}[t]
    \centering
    \includegraphics[width=0.40\textwidth]{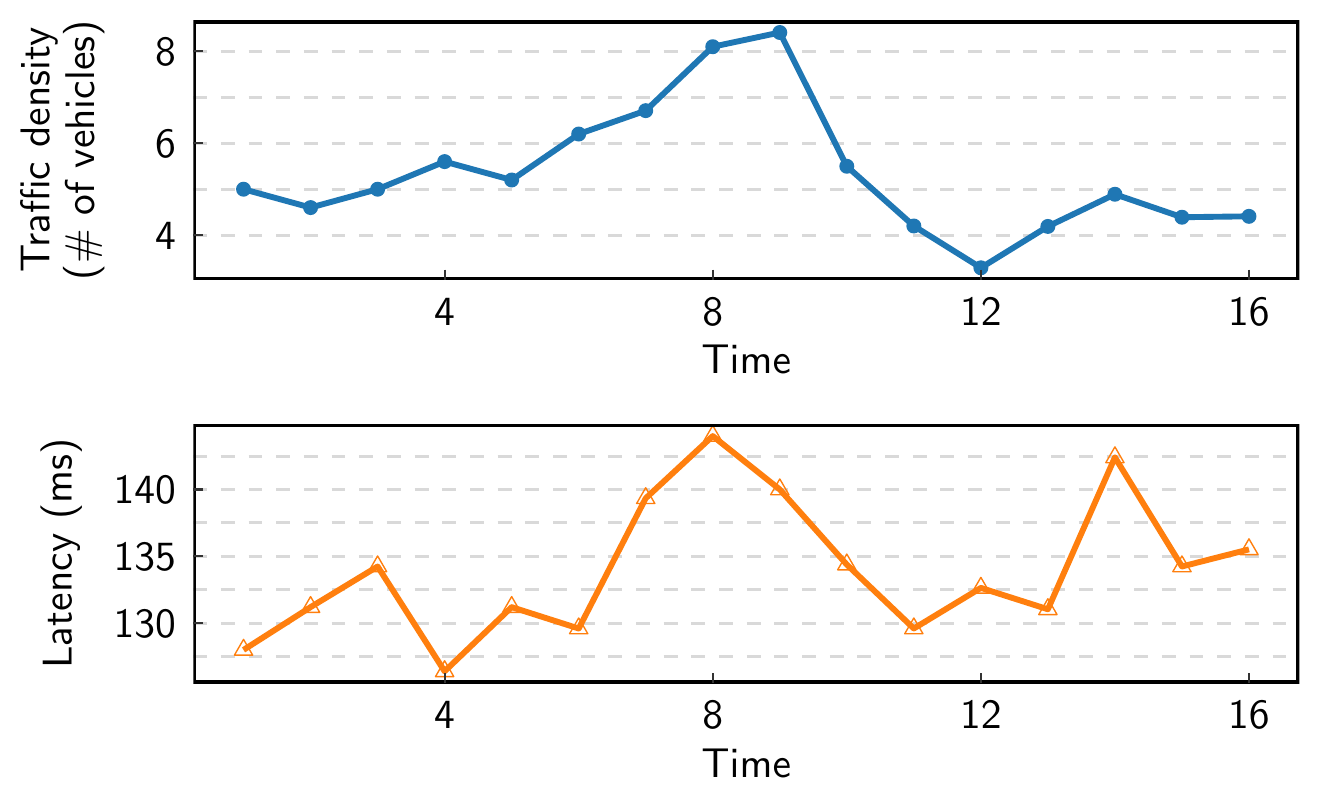}
    \vspace{-15pt}
    \caption{The end-to-end latency varies with different traffic densities in Apollo.}
    \vspace{-10pt}
    \label{fig:tf_density}
\end{figure}

\para{The Impact of Traffic.}
Traffic has a major impact on system latency.
We measure the end-to-end latency of Baidu Apollo's computation pipeline under various traffic densities.
The traffic density is defined as the average vehicle numbers in the scenario, within a radius of 25 meters. 
As shown in Figure~\ref{fig:tf_density},  traffic density has a strong correlation to the end-to-end computation latency in AV frameworks.
This observation motivates us to conduct a more detailed analysis of traffic impact on AV algorithms (Section~\ref{section:source}).
The variation brought by traffic also compounds the challenges of guaranteeing timely decision-making in complex driving scenarios.

\zixuan{Here you finished introducing a lot of background knowledge. Then how do they motivate your project? You should add one more subsection here summarizing key findings from the background knowledge, and your motivation for this paper.}

\section{AV System Latency Characterization}
\label{section:source}

%
System latency, according to our experiment results, has a strong correlation with driving safety.
To investigate and mitigate its impact on safety, we first propose a performance model and use it to conduct a detailed analysis of the system latency in state-of-the-art AV systems---including Baidu Apollo~\cite{apollo} and Autoware~\cite{10.1109/ICCPS.2018.00035}--- and representative AV algorithms~\cite{mmcv,Gog2021PylotAM,DBLP:journals/corr/abs-1808-10703,wu2019detectron2}.
Throughout our analysis, we observe that tail latency is the key factor in AV driving safety, and we propose corresponding mitigations in \secref{section:design}.


\subsection{Modeling AV Reaction Time}

\begin{figure}[]
    \centering
    \vspace{-30pt}
    \includegraphics[width=\linewidth]{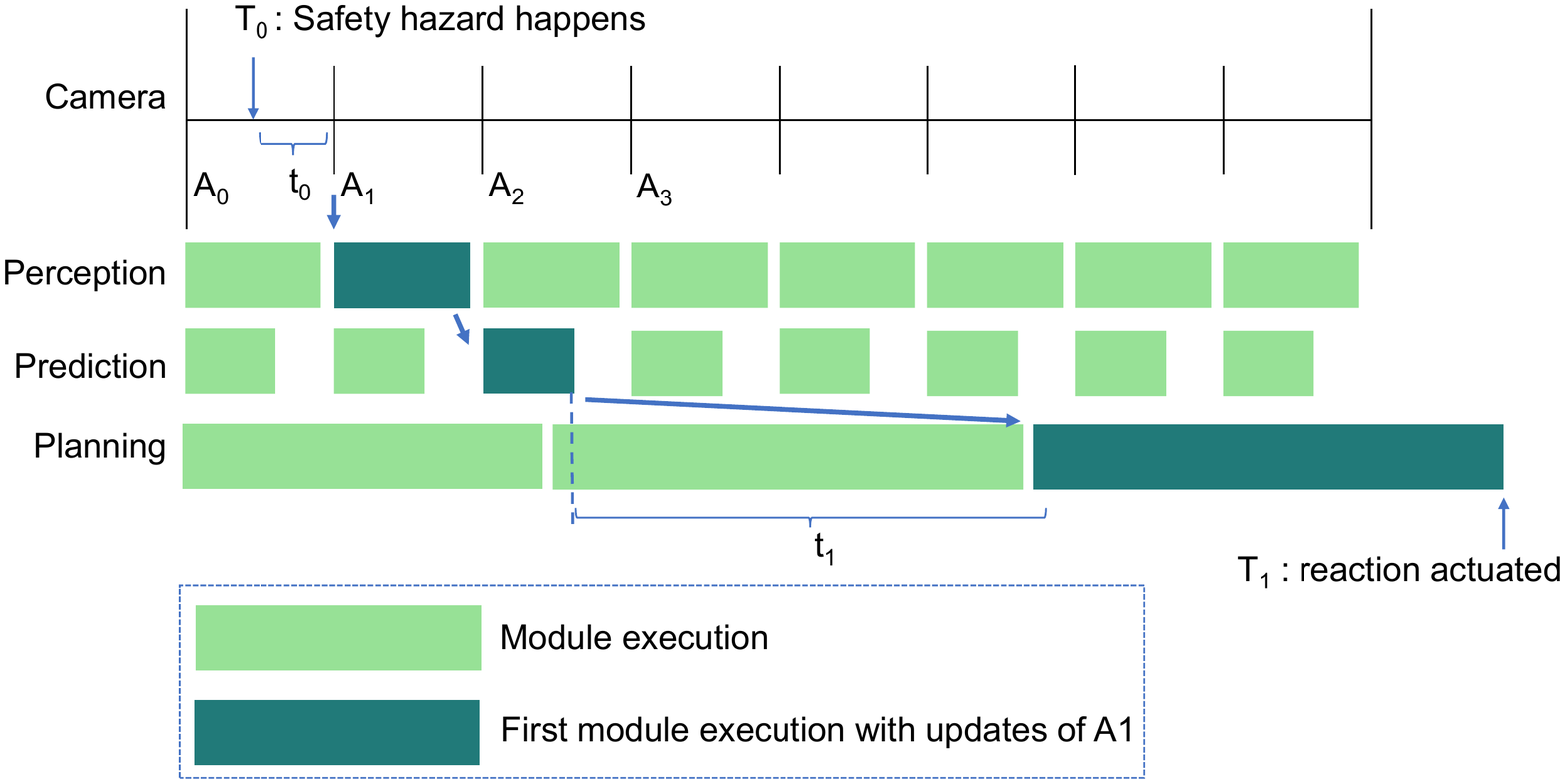}
    \vspace{-50pt}
    \caption{Reaction model in an example AV pipeline. $T_0$ is the time when a vehicle recklessly cut-in and may cause an accident if no action is taken; $T_1$ is the time when the AV makes a reaction to avoid the crash. The \emph{reaction time} in this example is the time between $T_0$ and $T_1$.}
    \vspace{-10pt}
    \label{fig:reaction}
\end{figure}



%
The Level-4 AV systems adopt the dataflow computation framework~\cite{Carbone2015ApacheFS,180560,199317} that organizes computation operations in a directed acyclic graph (DAG), where graph nodes represent computation operations and graph edges represent data passing operations.
In practice, to improve the system performance, these DAG operations are organized as a pipeline where computation can execute in parallel if they do not have dependencies.
%
%
In such a framework, data is periodically streamed from sensors into the DAG, processed by computations to reach to a final driving decision (e.g., vehicle acceleration) which is then sent to the vehicle control unit.
To ensure driving safety in such systems, DAG must process the sensor data and make the driving decision within a reasonable time.
According to prior reports~\cite{10.1145/3296957.3173191}, AV systems are expected to achieve much faster reactions~\cite{10.1145/3296957.3173191} (typically <150 ms) than human drivers~\cite{aaareaction} (typically 700-1500 ms).

To model the AV system-level latency, we first define the \emph{reaction time} as the time from a specific event (e.g., cut-in or deceleration) to the AV driving decisions.
%
%
We use this reaction time to model the AV system's pipeline computation latency.

\para{Modeling Reaction Time.} 
%
We first study major factors contributing to AV reaction time through a critical-path analysis in AV computation pipelines.
%
\figref{fig:reaction} shows an example of critical-path analysis based on camera sensor data processing:
The computation pipeline incorporates four stages, camera, perception, prediction, and planning. 
The camera sensor streams data into the AV pipeline at 30 Hz; and $A_{0}$, $A_{1}$ ... $A_{n}$ each indicates a single sensor frame.
%
%
Each colored node represents a computing process within an AV module, and the width of the node represents its execution time.
In this example, a safety hazard---a recklessly cut-in vehicle---happens at $T_{0}$.
This hazard is captured by the sensor which then streams the data into the pipeline as part of the $A_{1}$ frame.
Then, the perception module detects the abnormal speed and position of the cut-in vehicle, and passes this information to a prediction module that predicts the cut-in vehicle's behavior.
Based on the prediction result, the planning module eventually decides to slow down the vehicle to avoid the crash.
%

%
%
In the example, we consider the AV reaction time $t_{reaction}$ as the time from $T_{0}$ to $T_{1}$ and it consists of three parts:
(1) $t_{sensor}$ (shown as $t_{0}$ in the example): the time from the hazard to the next sensor frame generated;
its upper bound is determined by sensor frequency configurations;
(2) $t_{module}$: the accumulated latency from each computation module, i.e., perception, prediction, and planning;
%
(3) $t_{bubble}$ (shown as $t_1$ in the example): the bubble time when computation of the current frame is stalled, e.g., by the on-going computation.
%
%
Therefore, we describe reaction time by the following formula:
\vspace{-1em}

\begin{align}
  t_{reaction} = t_{sensor} + t_{module} + t_{bubble}
  \label{equation:reaction}
\end{align}

In the following section, we analyze the $t_{bubble}$ in \secref{section:bubble} and $t_{module}$ in \secref{section:tailchar}.

%

\subsection{Characterizing Bubbles}
\label{section:bubble}

We found the major sources of the bubble effects include the execution patterns, the entangled priority issue, and the limitations of end-to-end latency.

\para{Execution Patterns.}
An execution pattern refers to a triggering condition of the computation modules.
There are two types of execution patterns, timing-based (data is processed at a given frequency) and interrupt-based (data is processed upon arrival).
AV system designers choose different execution patterns based on module inputs and latency requirements.

Ideally, many computation modules should use the interrupt-based pattern to achieve the shortest reaction time.
While in practice, the system complexity can constrain the design choices and forces some modules to use timing-based patterns.
For instance, the fusion module synthesizes results from multiple perception modules, such as cameras (20-30 Hz) and LiDAR (10 Hz), to generate a single representation of the environment.
Ideally, the fusion module will use the interrupt-based execution pattern; but one of its downstream modules---the planning module---has an average latency at 100 ms scale.
Such a frequency-latency mismatch leads to a severe queuing effect between the fusion and the planning module, where planning cannot catch up and thus always operates on stale data.
As a consequence, AV frameworks such as Apollo~\cite{apollo} choose to run the fusion module at a lower frequency that matches the LiDAR frequency (10 Hz).
Such a design benefits the LiDAR sensors as there is no bubble time for LiDAR, whereas it degrades the reaction time for objects that are only detected in cameras.

\begin{figure}
    \centering
    \includegraphics[width=\linewidth]{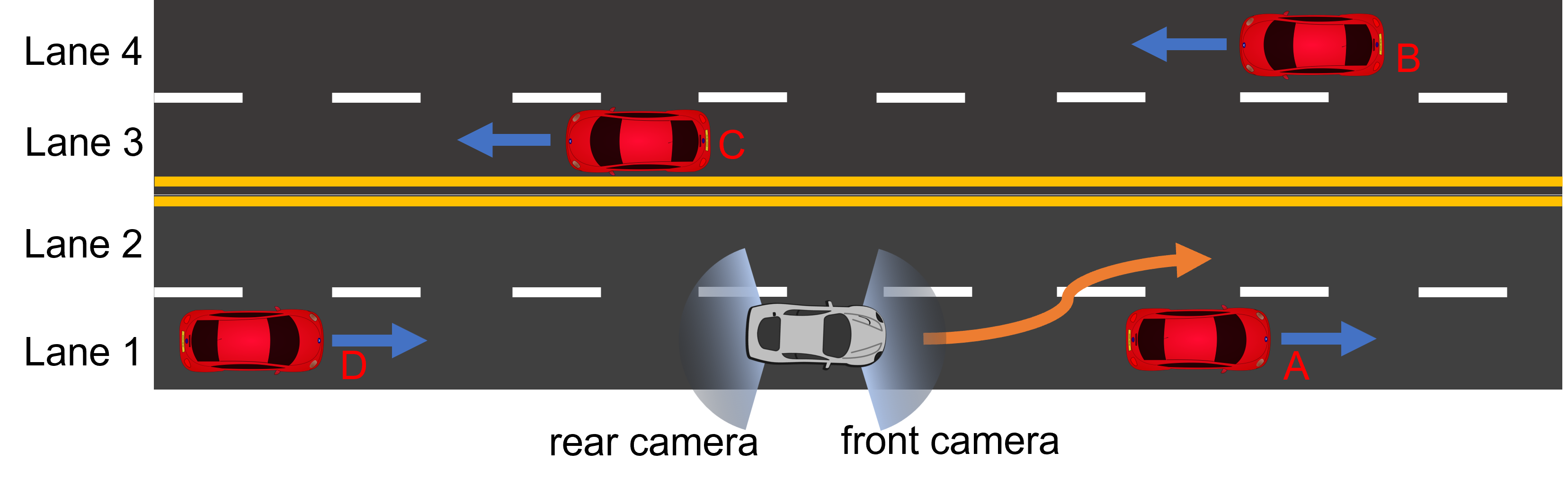}
        \vspace{-15pt}
    \caption{Entangled priority in current AV systems: 
    The front camera and rear camera are processed with equal priority although the front camera is more safety-critical;
    Vehicle A and B in the same sensor are also processed altogether despite different priorities.}
    \vspace{-10pt}
    \label{fig:reaction_model}
\end{figure}

We also found that statically assigned execution patterns, although widely used in AV frameworks, cannot adapt to the dynamic priorities at runtime:
For example, \figref{fig:reaction_model} shows a driving scenario, where the grey AV vehicle's rear camera can detect obstacles C and D while the front camera detects obstacles A and B.
In a static execution pattern, two sensors are assigned the same priority in the computation.
But in practice, obstacle A is more likely to cause safety issues, hence it should be prioritized.
In this example, AV should use a per-object priority to react to safety-critical issues faster.
But current AV systems, such as Apollo~\cite{apollo} and Autoware~\cite{10.1109/ICCPS.2018.00035}, do not support such fine-granularity control over the priorities.

\para{Observation 1:}
\emph{
A static execution pattern cannot achieve the ideal reaction time, as it prioritizes computations in a fixed manner without adapting to runtime requirements.
}

\haolan{Design of execution pattern in current AV framework -> concerns about Entangled priority}

\para{Entangled Priority.}
Current AV systems suffer from an \emph{entangled priority} issue, where objects (e.g., vehicles in the same camera data frame) are entangled with each other and thus share the same execution path in the pipeline, even if they have different priorities.
For example, the object detection module
delivers its results after processing all of the detected obstacles. 
Such an \textit{all-or-nothing} processing potentially hurts the reaction time in most near-collision scenarios where the AV only needs to prioritize one or two safety-critical obstacles~\cite{nhtsascenario}.
As shown in \figref{fig:reaction_model}, obstacles A and B in the front camera are processed at the same priority, while only obstacle A is more safety-critical.
Such a design motivates the need for \emph{partial updates} (\secref{subsec:adaptive-dataflow}) which assign priorities at an object level and process each object independently.

\para{Observation 2:}
\emph{
Current AV systems suffer from the entangled priority issue, which significantly degrades the vehicle's ability to make timely reactions in near-collision scenarios.
}

The above observation motivates an adaptive dataflow (\secref{section:design}) where object priorities are dynamically reconfigurable.
%

\para{Limitation of End-to-end Latency.}
%
End-to-end latency---defined as the time spent in the entire DAG processing---is widely used by prior works~\cite{10.1145/3296957.3173191} to model the vehicle reaction time.
However, the end-to-end latency only considers the computation latency $t_{module} + t_{bubble}$, while ignoring the sensor latency $t_{sensor}$ in Equation~\ref{equation:reaction}.

%
On the other hand, the end-to-end latency model assumes that the AV algorithms can instantly perceive the environment changes.
However, algorithms may take many frames to realize the environment changes in real-world driving scenarios, leading to more severe latency issues.
For instance, a prediction module may need several sensor frames to change the prediction of dangerous objects; and the fusion module needs several sensor frames to recognize a new object.
These issues indicate the limitation of using end-to-end latency to describe the vehicle reaction time, and motivate us to investigate module-level latency (\secref{section:tailchar}).

\subsection{Characterizing Module Latency}
\label{section:tailchar}

%
We analyze each module's latency in Baidu Apollo.
We further investigate the sources of tail latency and analyze their burst patterns with regard to driving scenarios.

\begin{figure}
    \centering
    \includegraphics[width=0.45\textwidth]{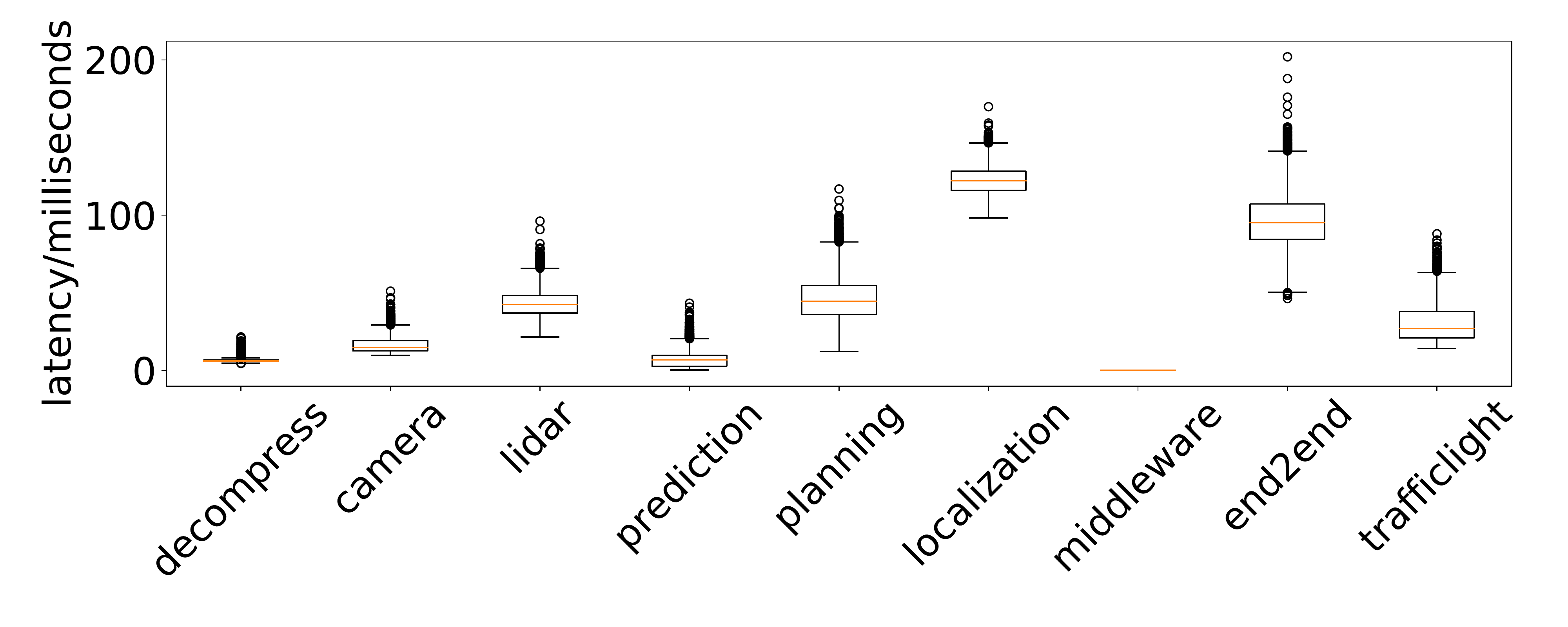}
     \vspace{-10pt}
    \caption{Latency breakdown of AV modules in Apollo. We only show the latency of the main camera and LiDAR.}
    \vspace{-15pt}
    \label{fig:all_breakdown}
\end{figure}

\figref{fig:all_breakdown} shows our latency analysis of major modules in Apollo systems.
\texttt{LiDAR} and \texttt{camera} are part of the perception module.
The \texttt{decompress} stands for the latency to decompress camera images.
The \texttt{end2end} indicates the end-to-end latency of AV pipelines from LiDAR sensor data to AV control actuation.
The \texttt{middleware} indicates time spent in the AV communication layer, such as cyberRT~\cite{cyberrt} and ROS~\cite{Quigley2009ROSAO}, whose latency is negligible.
\texttt{localization} and \texttt{traffic light} are not in the critical path of end-to-end latency, they are running asynchronously in a fixed time interval.

%
%

\para{Tracing Module Latency.}
%
%
We analyze the tail latency in AV frameworks, from two different aspects: algorithm and system.
To study the algorithm factors, we collect the input trace of each module and replay them in isolation to exclude system-level factors such as cache contention;
to study the system factors, we use the same input (traffic scenarios) under different system configurations, and measure the latency variation.
We also develop a lightweight tracing framework by time-stamping the input and output of each module.
%
%
%
%
With this setup, we study different modules' latency and corresponding causes.

\begin{figure*}
    \centering
    \begin{subfigure}[t]{.23\textwidth}
        \includegraphics[width=\linewidth]{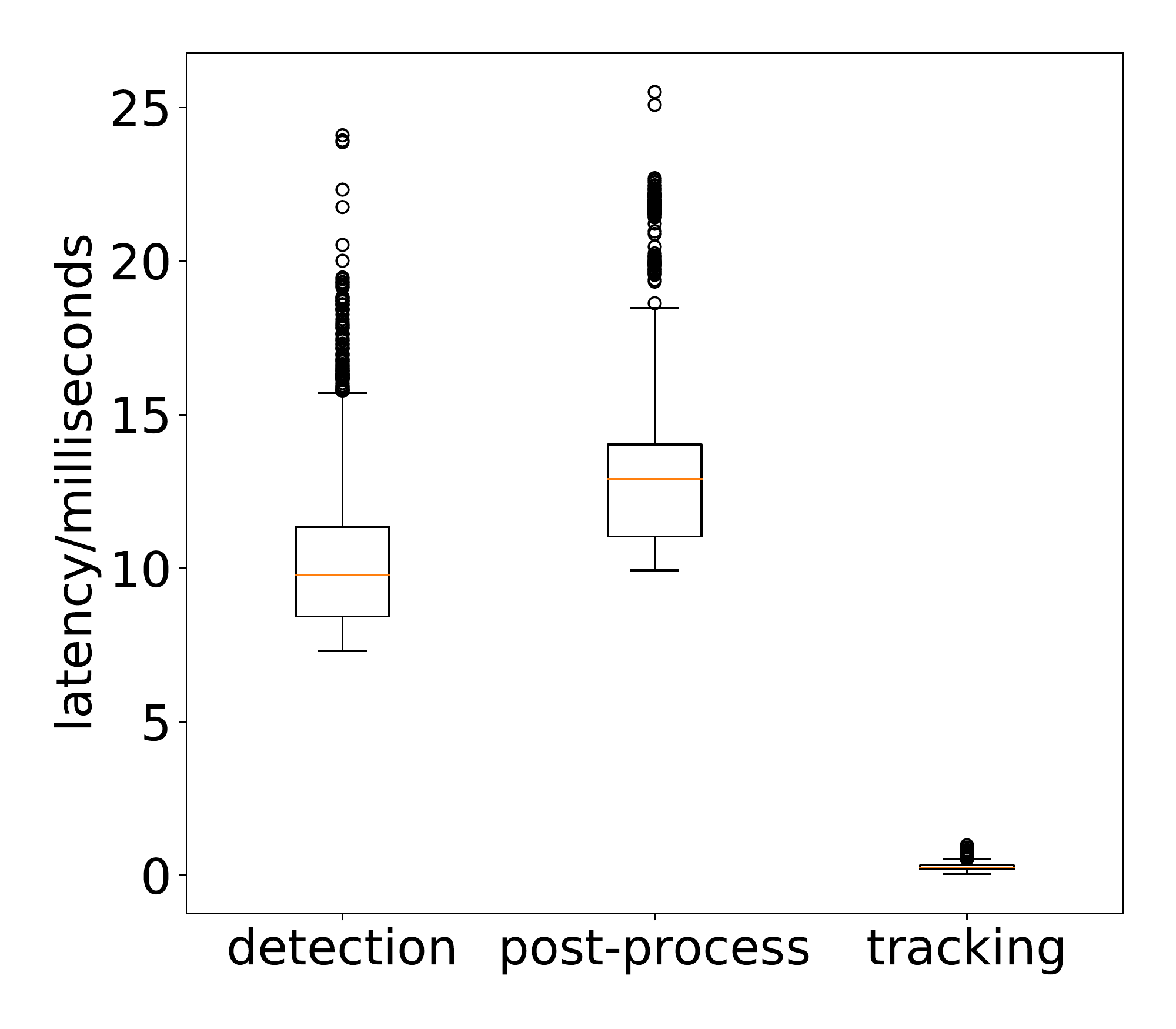}
        \caption{Latency breakdown of Apollo LiDAR perception modules.}
        \label{fig:perception-pipeline}
    \end{subfigure}
    \hfill
    \begin{subfigure}[t]{.23\textwidth}
        \includegraphics[width=\linewidth]{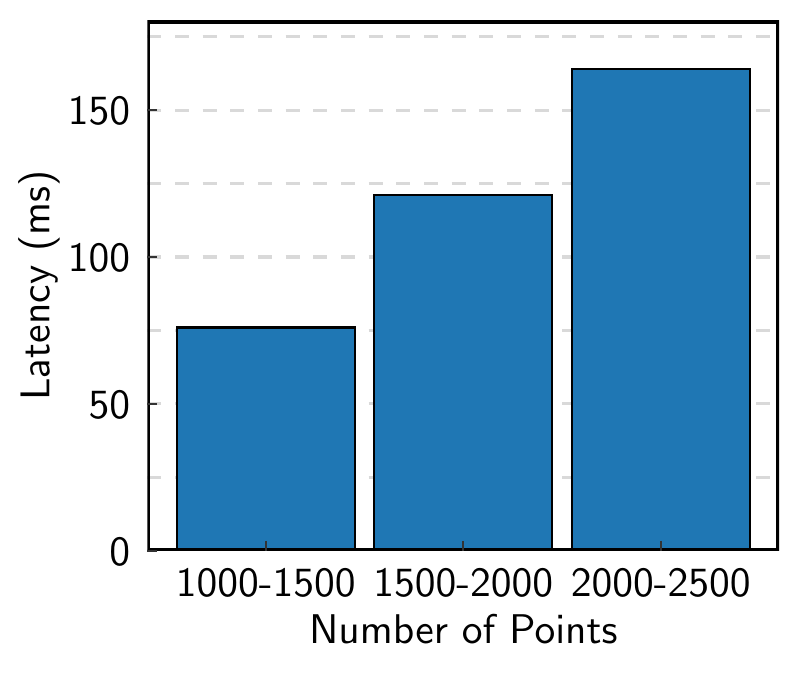}
        \caption{Performance of localization with varying numbers of LiDAR cloud.}
        \label{fig:local}
    \end{subfigure}
    \hfill
    \begin{subfigure}[t]{.23\textwidth}
        \includegraphics[width=\linewidth]{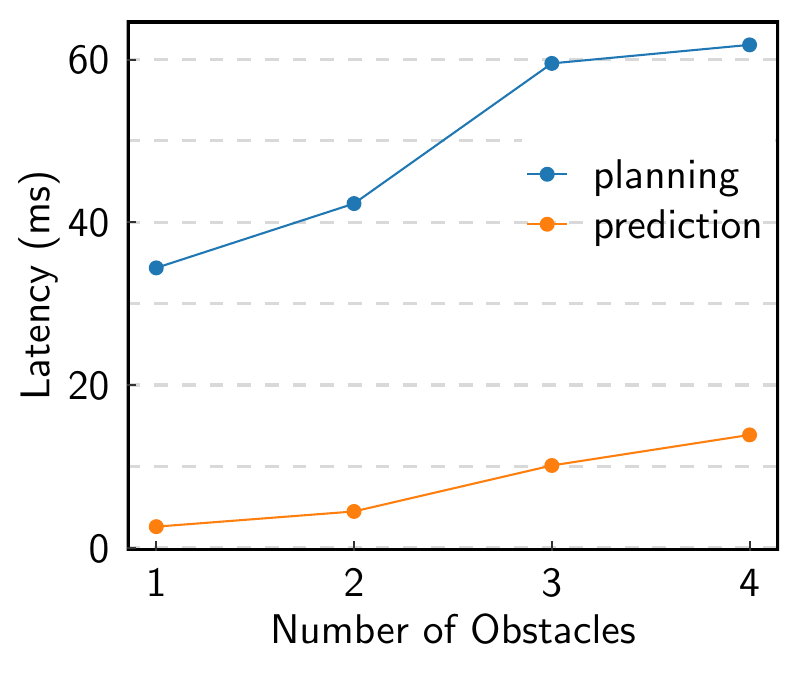}
        \caption{Performance of prediction/planning with varying safety-critical obstacle numbers. }
        \label{fig:pred_perf}
    \end{subfigure}
    \hfill
    \begin{subfigure}[t]{.23\textwidth}
        \includegraphics[width=\linewidth]{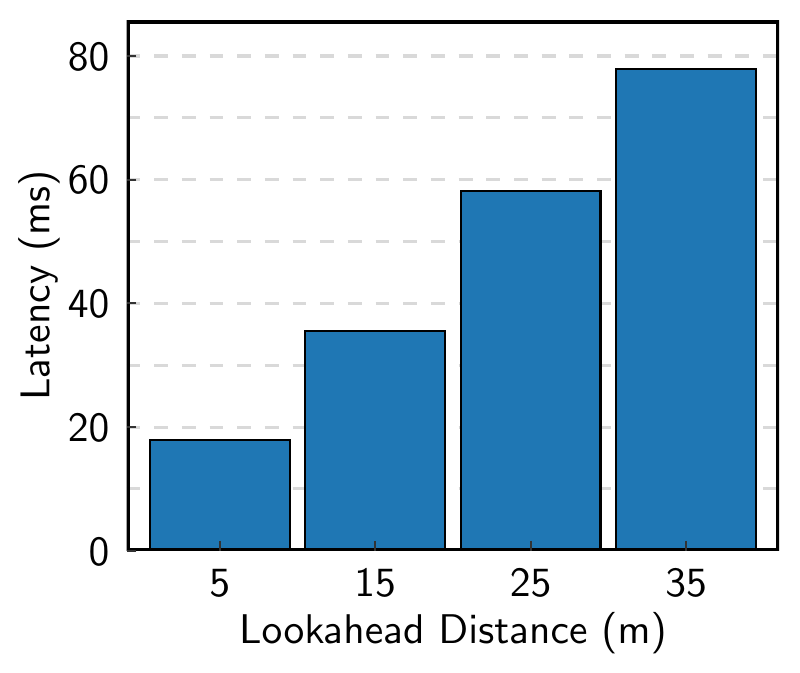}
        \caption{The latency increases with longer lookahead distance.}
        \label{fig:lookahead}
    \end{subfigure}
    \vspace{-10pt}
    \label{fig:perf-eval}
    \caption{Performance characterization of AV modules. (a) shows the latency breakdown of LiDAR perception pipeline. (b)(c) shows the bursty pattern of localization, prediction and planning with regard to obstacle numbers. (d) shows the level of being predictive also affects planning latency. \haolan{b,c combined}  }
    \vspace{-10pt}
\end{figure*}

\para{Perception.}
The AV typically employs multiple sensors with different positions and angles to avoid blind spots.
Each sensor streams data to a downstream perception module that extracts features such as object position and velocity.
These perception modules heavily rely on deep neural networks (DNNs) for segmentation, object detection, and object tracking.
%
%
These DNNs typically exhibit stable latency as their code execution is predictable and don't have conditional branches.
But other non-DNN components, such as the post-processing steps, impose more latency variations.
\figref{fig:perception-pipeline} shows the performance breakdown of the Apollo LiDAR pipeline, where \texttt{detection} and \texttt{post-processing} dominate the overall performance and they are not DNNs.
%
%
This motivates us to further investigate the source of these latency fluctuations, and as shown in  \secref{section:resource_contention}, the major causes are contention in CPU last-level-cache and GPU.
%

\para{Localization.}
Among all localization algorithms in Level-4 AV systems, the \emph{scan match} is the most popular one~\cite{DBLP:journals/corr/abs-1711-05805}.
%
This algorithm estimates the object position at centimeter-level accuracy by aligning the real-time detected point cloud with the offline collected point cloud (HD map).
This matching process is affected by the number of points in the point cloud, and according to our profiling result in Apollo (\figref{fig:local}), the localization module has higher latency when it needs to process more obstacles.
%

\para{Fusion.}
%
Level-4 AV systems adopt multiple redundant sensors including LiDAR, camera, and radar, to avoid single-point failure.
%
%
Each of these sensors sends data to the corresponding perception pipeline that streams results to a centralized fusion module.
Based on the perception results, the fusion module initializes and maintains the tracking records (including their speed, acceleration, and position) of the surrounding objects.
It also deletes the tracking record when the objects are out of sensor ranges.

Some designs in the fusion module can add to the reaction time.
For example, to initialize the tracking record for a newly detected object, the AV needs to prevent "ghost track" (initialize a non-existing object).
A typical statistical method is to initialize only when the perception pipeline detects the new objects $a$ times ($a$ is a predefined threshold) in the previous n frames.
Therefore, in the first $a$ frames, even if we detect the object, the downstream module (prediction and planning) will not process the object and make a reaction to its potentially dangerous behavior.
Such design negatively impacts the reaction time to newly detected objects.

\begin{figure}
    \centering
    \includegraphics[width=0.45\textwidth]{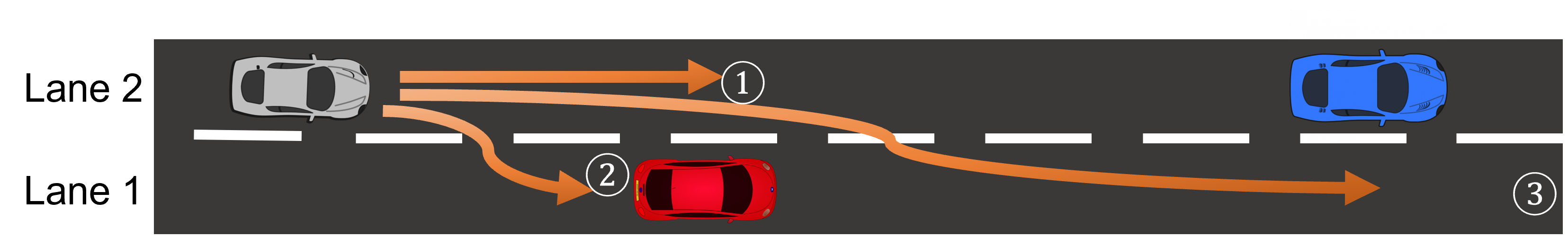}
    \caption{Three possible trajectories for planning modules.}
    \vspace{-10pt}
    \label{fig:scenario3}
\end{figure}

\para{Prediction \& Planning.} The prediction module predicts the future movement of dynamic objects (vehicles, pedestrians).
The planning module leverages such information to plan a crash-free and comfortable driving trajectory.
Fast response is necessary for planning modules as it reflects the AVs' capability to react to real-world hazards.
Prediction and planning modules also exhibit burstiness in varying traffic, as both modules need to process more obstacles in heavier traffic~\cite{fan2018baidu}.
\figref{fig:pred_perf} shows the latency of prediction and planning with respect to safety-critical object numbers (non-safety-critical objects have fewer impacts on the latency). We observe heavy traffic can simultaneously lead to 1.5$\times$ and 1.8$\times$ overheads in prediction and planning latency.



\para{Observation 3:} 
\emph{
Traffic has a major impact on system latency:
the number, category, and distribution of obstacles, and the road condition will yield different performance variations.
Worse even, the latency requirements typically become more stringent
in scenarios with heavy traffic.
}

\vspace{2pt}


The planning module takes a hierarchical structure: The behavior planning layer determines AV behaviors including lane following or change to the right lanes~\cite{6856582}.
Based on that, the motion planning layer will compute the actual trajectory.
AV driving intentions can also affect latency. 
In \figref{fig:scenario3}, the \ding{172} and \ding{173} depicts the two trajectories produced by following lanes or changing lanes.

The planning module is \emph{predictive}, which means if the environments evolve as it predicts, the AV is safe within a time window.
The AVs are guaranteed to avoid crashes if: (1) No new obstacles are detected.
(2) the dynamic objects are moving as predicted.
Forecasting and decision-making in the long term are harder, as it involves more uncertainties.
For example, motion planners need to determine a \emph{lookahead distance}: the motion planners will only consider a trajectory within the length of the lookahead distance. With longer lookahead distances, the planning module will be more "foresighted": it can compute more accurate paths, but it usually takes longer.
With shorter distances, the decision may be suboptimal, but faster~\cite{frenettrajectory}.
Figure~\ref{fig:scenario3} shows a case that the \ding{174} is only possible with a longer lookahead distance.
Figure~\ref{fig:lookahead} shows the planning latency with varying lookahead distances.
Such a tradeoff is also common in the prediction module. To predict long-term behavior (e.g. future movement in 10 seconds rather than 2 seconds), prediction modules commonly leverage DNNs with bigger capacity (usually longer latency)~\cite{DBLP:journals/corr/abs-1808-05819}.

\haolan{Use the latency difference between longer time window prediction and shorter prediction.}
\para{Observation 4:}
\emph{
Software modules, such as prediction and planning, compute predicatively in the critical path, which may hinder AV from fast reaction.
}

\para{The Latency-Accuracy Tradeoff.}
AV systems need to make a tradeoff decision between latency and accuracy, including network architecture, sensor resolution, and module parameters (like the lookahead distance)~\cite{Gog2021PylotAM}.
Many of those decisions are pre-configured and fixed at runtime, which is also hard for AV designers to find the optimal configuration.
The fixed design is not suitable for Level-4 AV systems with dynamic latency requirements.

\para{Observation 5:} Most of the current AV algorithms have fixed configurations and performance, whereas AV systems need to dynamically adjust their latency requirements based on different driving scenarios. Such a mismatch hinders AVs from quickly reacting to potential safety hazards.

\para{Safety Redundancy.} Many machine learning algorithms for AVs are susceptible to long tail problems~\cite{DBLP:journals/corr/abs-2107-08142}. To this end, AV frameworks adopt safety redundancy mechanisms, in case the regular algorithms no longer work. One example is uncertainty calibration: when a DNN is not confident with its prediction,
 the AV will use a more powerful downstream calibration module. Such a calibration process adds latency to the critical path when the upstream module is not confident, therefore slowing down the reaction time. The \texttt{segmentation} in \figref{fig:bursty} is an example of calibration, we can see it adds to about 28 ms in the critical path at most.





\subsection{System-Level Resource Contention}
\label{section:resource_contention}

In the following, we investigate 
the AV system hardware resource metrics and quantify their impact on latency variation.

\para{Last Level Cache.}
The last level cache (LLC) in CPU cache hierarchy has a direct impact on the performance, especially for memory-intensive tasks~\cite{10.1145/3302424.3303977}.
Some AV tasks, such as segmentation, process a big camera image or point cloud sets. Therefore, the cache miss rate will have a major impact on those tasks.
To quantify the contention on LLC, we employ the PAPI library~\cite{icl:16} to investigate the microarchitecture characteristics, such as cache miss and branch misprediction rate. PAPI collects the data from the performance monitor counter (PMC) at a low runtime overhead. To exclude the variation introduced by different inputs, we replay the same data input collected by our microbenchmark.

\begin{figure*}[t]
    \centering
    \begin{subfigure}[t]{.19\textwidth}
        \includegraphics[width=\linewidth]{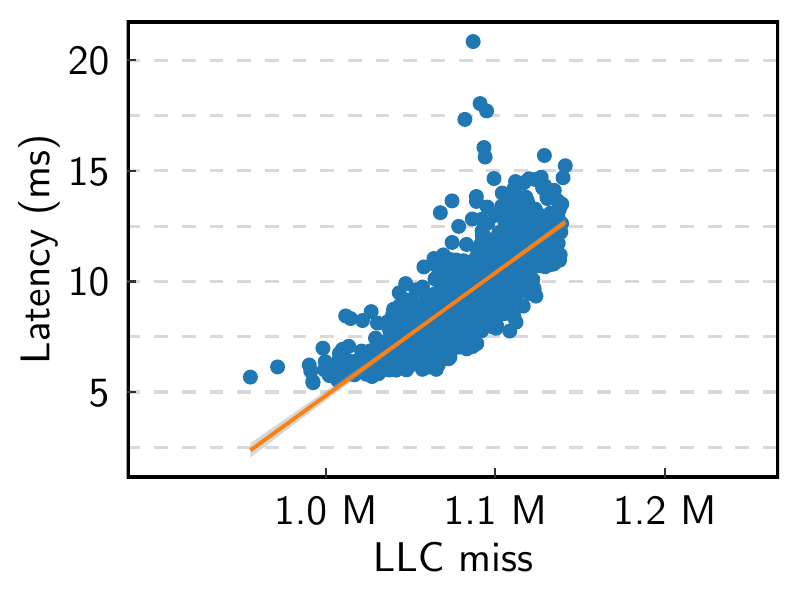}
        \caption{The relation between LLC misses and latency.}
        \label{fig:cache}
    \end{subfigure}
    \hfill
    \begin{subfigure}[t]{.19\textwidth}
        \includegraphics[width=\linewidth]{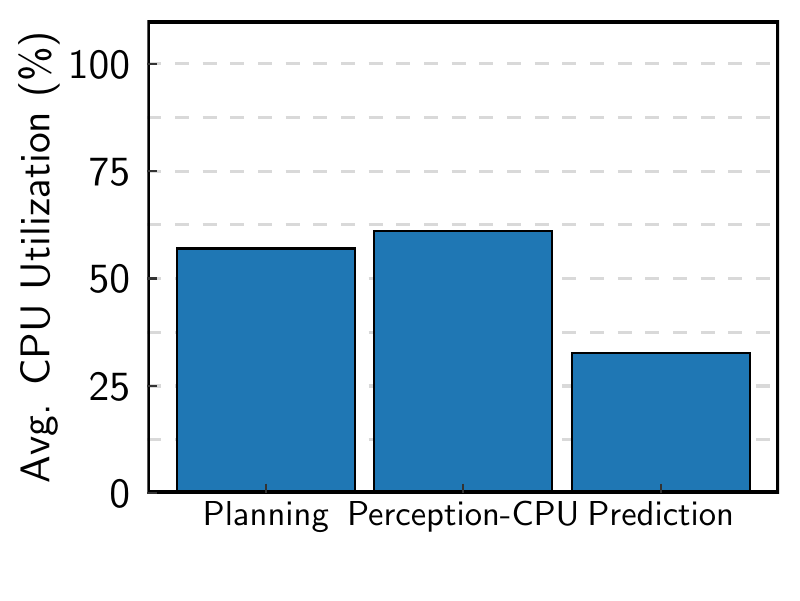}
        \caption{CPU utilization of bursty modules.}
        \label{fig:util}
    \end{subfigure}
    \hfill
    \begin{subfigure}[t]{.19\textwidth}
        \includegraphics[width=\linewidth]{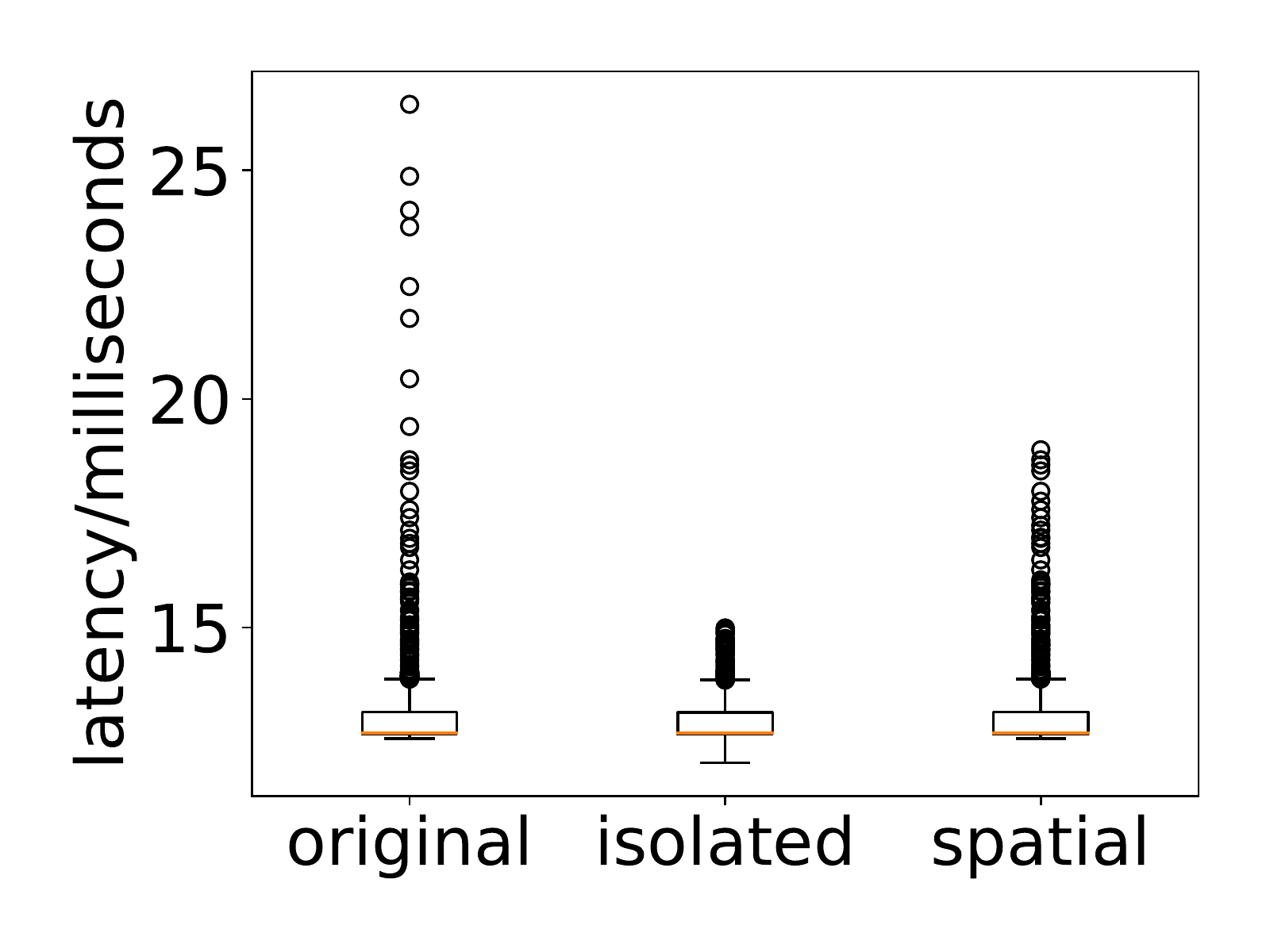}
        \caption{GPU kernels' latency under varied policies.}
        \label{fig:gpuiso}
    \end{subfigure}
    \hfill
    \begin{subfigure}[t]{.19\textwidth}
        \includegraphics[width=\linewidth]{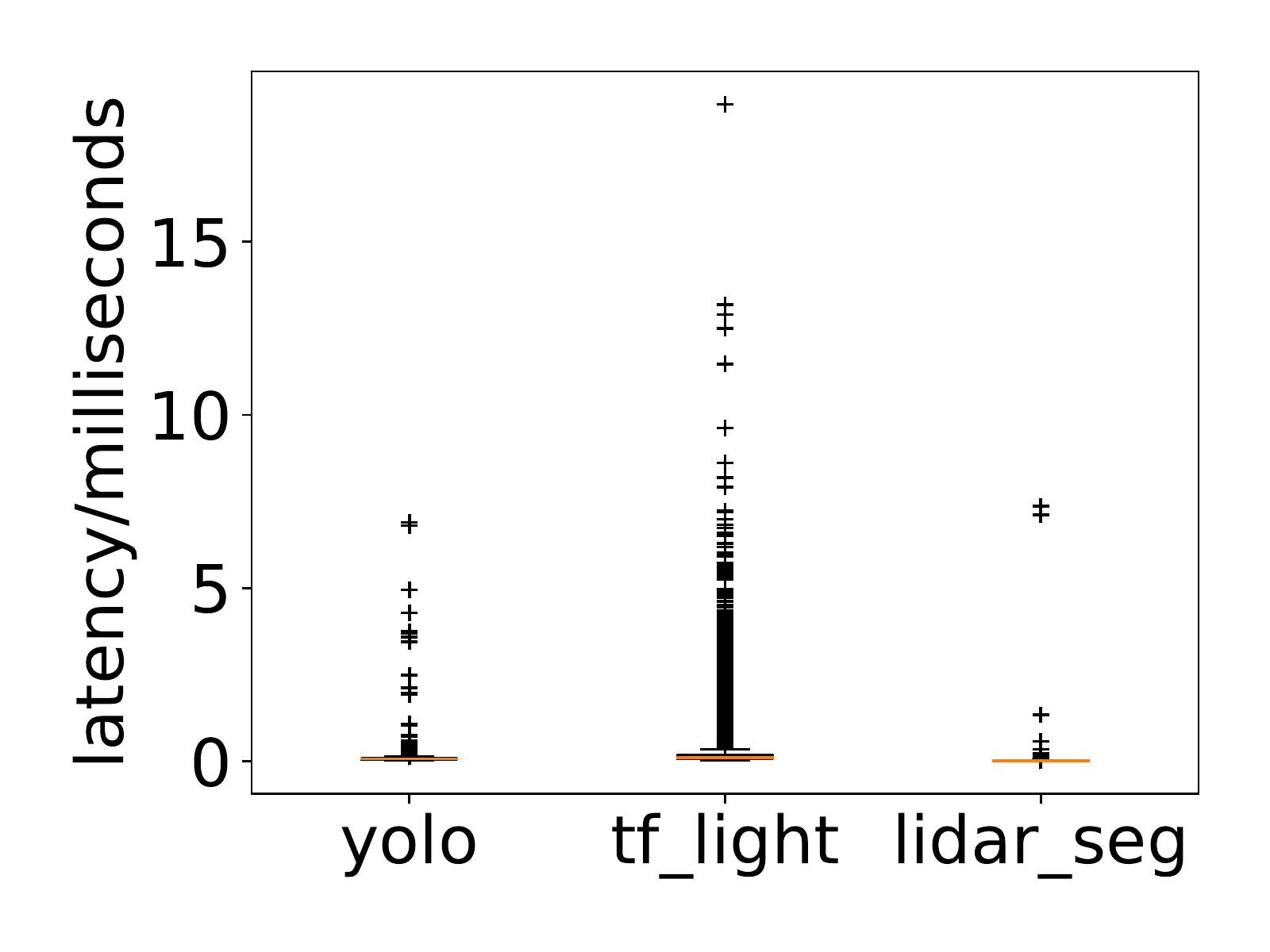}
        \caption{Latency of transferring data to GPU.}
        \label{fig:memcpy}
    \end{subfigure}
    \hfill
    \begin{subfigure}[t]{.19\textwidth}
        \includegraphics[width=\linewidth]{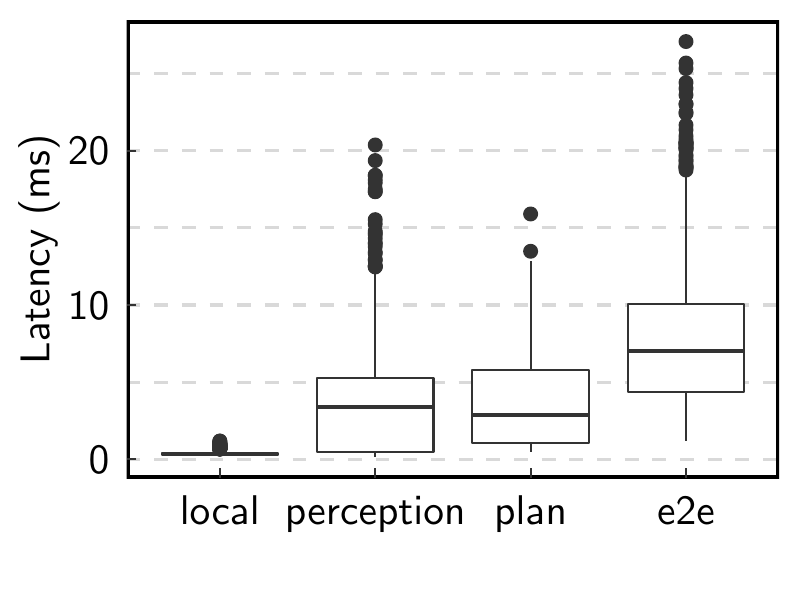}
        \caption{Latency breakdown of prediction pipelines.}
        \label{fig:pred_overlap}
    \end{subfigure}
    \vspace{-10pt}
    \caption{Latency and utilization characteristics in AV modules.}
    \label{fig:perf-latency}
    \vspace{-10pt}
\end{figure*}

We select the object clustering algorithm, a widely-used algorithm in LiDAR perception pipelines.
\figref{fig:cache} presents the scatter point and the fitted linear model, which shows
a strong correlation between LLC miss and latency variation. The slope is around 85.36. As the access latency is approximately 100 ns in local DRAM, the theoretical variation brought by the LLC miss matches the profiled latency variation.
We also profile other microarchitecture states that influence latency results such as \textit{branch misprediction} and \textit{TLB miss}.
We find they only bring several hundred microseconds variation, which has trivial impacts on the latency.

Similarly, a substantial set of AV algorithms work on
large in-memory data structures, including high-resolution images, point clouds, and feature maps produced by the DNN.

\para{Observation 6:} 
\emph{
The memory-intensive workloads will cause serious resource contention in the LLC in AV systems. The AV system needs to carefully manage such contentions in co-located tasks.
}

\para{Scheduling.}
The resource scheduling layer (middleware) of AV systems, such as cyberRT and ROS, needs to allocate CPU to different tasks and guarantee their timely response~\cite{apollo,Quigley2009ROSAO}. 
Due to the strict real-time requirements, current AV systems allocate resources statically.
For instance, the prediction module in Apollo maintains a thread pool with fixed number of threads to avoid performance interference.
In \figref{fig:util}, we measure the CPU utilization of three CPU-intensive tasks in Apollo: prediction, planning, and the segmentation calibration of perception modules. 

This observation also applies to hardware accelerators such as GPU.
Today's AV framework does not spatially share the GPU, by serializing the concurrent computing process.
However, such exclusive access leads to queuing. e.g., AV may launch two kernels in the same GPU at nearly the same time, then the latter one has to wait for the completion of the other.

\figref{fig:gpuiso} shows the DNN inference latency in Apollo's LiDAR segmentation modules, the \texttt{original}, \texttt{isolated}, and \texttt{spatial} respectively indicate latency profile onboard, on an isolated GPU, and with spatial concurrent execution.
The queuing also impacts the PCIe data transferring. \figref{fig:memcpy} shows a long tail latency in three GPU tasks. The worst-case (18.96 ms) could be 40 $\times$ the mean latency (0.466 ms).

\para{Observation 7:} 
\emph{
Current AV middleware design (CyberRT, ROS2, \textit{etc}.) employs static computation resource allocations
based on AV computation graphs. This leads to low utilization and system throughput with bursty AV algorithms.
}

\section{Tail Latency Mitigation}
\label{section:design}

Motivated by the observation (\textbf{Observation 1-7}), this section mainly summarizes and discusses design hints (\textbf{H1-H5}) for Level-4 AVs systems to mitigate the tail latency.






\subsection{Adaptive DataFlow}
\label{subsec:adaptive-dataflow}

Our observation (\textbf{Observation 1,2,4,5}) shows that current AV frameworks present fixed dataflow, which hinders AV from reacting fast in near-collision scenarios. 
This motivates us to reconfigure the AV dataflow dynamically at runtime, in favor of flexible and reliable latency control.
Specifically, we propose the following design hints (\textbf{H1-3}) to make AV systems adaptive in different driving scenarios.

\hint{H1: }{Supporting partial updates.}

\hint{H2: }{Prioritizing safety-critical information in the pipeline.}

\hint{H3: }{AV algorithms should adjust latency upon demand.}

\haolan{Partial Update, while the least updates will propagate to the motion planning later. Motion planning module can perform crash avoidance checking to decide if a replanning is needed.}

\begin{figure}
   \includegraphics[width=0.45\textwidth]{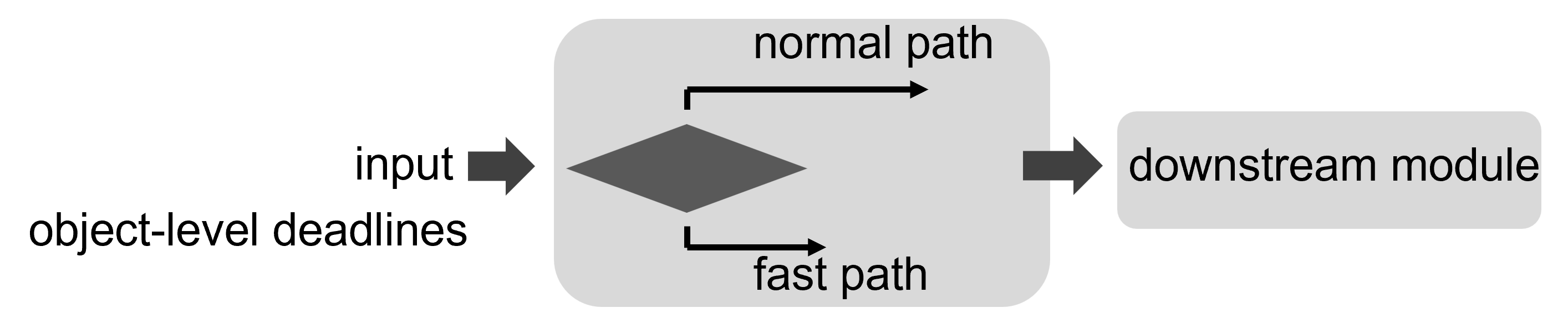}
   \vspace{-10pt}
    \caption{
   Before execution, the AV module will determine whether to take the fastpath or the normal path based on the remaining time budget. }
   \vspace{-10pt}
    \label{fig:fastpath}
\end{figure}

\para{Fastpath.}
To prioritize the dataflow of safety-critical information, we propose and implement the \textit{fastpath} mechanism.
As shown in Figure \ref{fig:fastpath}, 
when the module executes, it will determine between the normal path and the fastpath, with the remaining time budget, which is derived from the deadlines and the time already consumed in upstream processing. We propose an \emph{object-level} deadline to compute the priority of dataflow, to inform fastpath mechanism about which upstream data to prioritize.

\para{Object-level Deadlines.} 
To track the time budget, AV frameworks such as Apollo, Autoware, and D3 attempts to specify the deadlines on a \emph{system-level} (end-to-end deadlines) and \emph{module-level} (module deadlines)~\cite{10.1145/3492321.3519576,10.1145/3296957.3173191,Luo2019TimeCA}. 
Our Observation 1 and 2 shows that such specification entangles the priorities of different sensors and objects.
It is easier and also meaningful to specify a fine-grained \emph{object-level} deadline based on their stringency. We can derive the object deadlines based on responsibility-sensitive safety
~\cite{ShalevShwartz2017OnAF} as we introduce in Section~\ref{section:latencyrequirements}. 
Based on the object deadlines, we can compute the deadline of each message as the earliest deadline among all the objects detected in this sensor frame.
In the following, we introduce two approaches to build fastpath in AV modules.

\para{Fastpath in Prediction.}
The prediction module will sequentially process the surrounding obstacles and predict their future movement, which is later incorporated in planning modules to avoid crashes~\cite{xu2020data}. Processing all the obstacles helps AVs to find the optimum decision. However, when the time constraint violation is coming up, we focus on the most important obstacle and try to avoid crashes with that obstacle.
Based on the distance from the AV, we first sort out the obstacle within 20 meters and process those partial inputs to quickly unblock the downstream module. 
The remaining information will be later propagated to the downstream module.



\para{Fastpath in Planning.} 
Inspired by Observation 4, we can prune the computation for long-term planning to achieve a faster reaction time. 
For example, 
we can choose a less powerful prediction module with a shorter time window, along with a planning module with a shorter lookahead distance.
In this way, the fastpath will react faster to the dangerous obstacles, but it may generate suboptimal trajectories, in terms of commuting efficiency.

\para{Tradeoff.} 
Fastpath trades algorithm optimality for faster response time. For example, our fastpath design in prediction and planning sacrifices the long-term algorithm capability (trajectory optimization), while retaining short-term capabilities (crash avoidance).
Therefore, such a design won't bring safety issues, yet it can hurt driving experiences and traveling efficiency.
However, building fastpath in the perception module may miss some safety-critical objects. As a result, COLA only builds fastpaths in prediction and planning.

Another potential downside of the fastpath is that it may lead to more frequent execution of downstream modules, which can further lead to queuing. To avoid that, our implementation starts an extra worker to run the downstream computation, which only applied to stateless downstream modules. Another option is to build a fastpath also for the downstream module, so that they can also support more frequent execution.

\para{Comparison With Emergency Control.}
Current AVs have an emergency override mechanism: when the vehicle detects some emergency scenarios (e.g. obstacles distance < 1m ), the AV will perform an emergency stop to avoid crashes. For emergency control, the vehicle can only perform a limited set of choices including a sudden stop or steer. In addition, frequent emergency control greatly hurts the AV driving experience.
Note that fastpath is trading the driving efficiency for faster reaction to potential hazards, which is different from the hierarchical emergency control.

\subsection{Proactive Processing}

\haolan{Early cancellation, criticality-aware.}

\hint{Hint 4: }{Proactively performing computation can alleviate the latency variation.}

\para{Ahead-of-time Processing.}
In Apollo, the prediction module has multiple inputs: the vehicle position from the localization module, the obstacle and road information from the perception module, and the planned trajectory information from the planning module. The prediction modules are triggered by the arrival of perception output, then firstly it will fetch the newest localization and planning frames and update its state.
Figure \ref{fig:pred_overlap} shows the performance breakdown of the prediction modules. \texttt{e2e} indicates end to end latency of the prediction module. \texttt{local}, \texttt{perception}, \texttt{planning} indicate processing incoming messages from the localization, perception, and planning module.
As those computations don't have any dependency, they can be performed independently ahead of time: when the planning module delivers its result, the AV framework can start a thread to update the state in prediction.
In this way, the prediction module can directly fetch the results during execution.

Another example is that we find the planning module relies on the driving intention from the behavior planning layer (such as changing to the left/right lanes). We can proactively compute the trajectory to change lanes even when the current instruction is to follow the original lanes. To this end, when the instruction changes we may directly use the proactive results.
Proactive processing trades extra computation for faster reaction. We can improve it by early cancellation: stop the proactive computation when the current processing becomes invalid or can be overridden by new execution. E.g. when the AV starts to change to the right lanes, proactively computing changing to the left lanes is unnecessary.

\subsection{System-Level Optimization}\label{section:system}

\hint{Hint 5: }{AV algorithms should improve their predictability and bursty pattern during heavy traffic. Such information can be used by the runtime and the OS for better scheduling. }


\para{Best-effort work stealing.}
Based on Observation 5, CPU resource is quite underutilized due to static specification. As infrequent preemption overhead is negligible for millisecond-level AV workloads, utilizing the underutilized resources to accelerate AV workloads is possible. To this end, we propose \textit{best-effort work stealing} mechanism and implement it in Baidu Apollo. The gist of this idea is stealing CPU resources without imposing performance violations on other tasks. We prohibit work stealing if it may lead to violations.

In Apollo, tasks are wrapped as coroutines and are scheduled by the cyberRT runtime system. To achieve good performance isolation, cyberRT will pin tasks to a processor group. The cyberRT at runtime schedules those coroutines based on predefined priority. To guarantee real-time performance, each CPU group is set to accommodate the maximum CPU requirements of all its tasks. 

We implement \textit{nice guest} coroutines that can conditionally migrate to other processor groups: only when such migration won't bring time violation to other tasks. 
To determine if the host processor group can serve the nice guest coroutine without performance violation, we observe that the prediction module adopts different code paths based on the number and type of obstacles, and process them in order.
Therefore, we propose a linear predictor to estimate the latency bound of prediction modules, with n category (vehicle, people, cyclist, \textit{etc}). $Time_{i}$ and Offset is set by extensive profiling.
\begin{align}
  {Latency} =  {\sum_{i=1}^{n}{(Time_{i} \times Number_{i})}} + {Offset}
\end{align}

\vspace{-8 pt}


\haolan{delete for now}

\vspace{-3pt}

\section{Evaluation}

This section evaluates the performance and safety of the design techniques in Section \ref{section:design}.

\subsection{Experimental Methodology}

\para{Experimental Setup.}
Due to the dangerous nature of our experiments,
we use a high-fidelity simulator, Carla~\cite{Dosovitskiy17} to simulate the environments. 
Previous research conducts extensive comparison experiments to verify the fidelity of safety testing in the Carla simulator~\cite{Fremont2020FormalST}.
We run real industrial AV software and hardware along with the simulator (We list the specific details in Appendix~\ref{appendix:simulation}).


As the Carla simulator consumes lots of CPU/GPU and memory. To avoid possible performance interference,
we run the simulator on another machine with GTX 1080Ti, connected with the host AV machine via 1 GB/s Ethernet cable.

\para{System Configuration.} We run our experiment on a server system, which is also a common practice in the industry such as Waymo and Pony.ai~\cite{apollo,zhao2019towards,waymo,avhardware}. Our machine has a 12-cores Intel it-8700K CPU with 32GB DRAM and 256GB NVMe SSD. The GPU devices include 2 NVIDIA Titan-Xp with 12GB device memory. The operating system is Ubuntu 18.04, with an Apollo customized real time patch~\cite{apollortkernel}, which is the same configuration as Baidu Apollo's on-vehicle testing~\cite{apollo}.
We also disable the CPU frequency scaling to avoid its impact on latency.

\para{Baselines.}
We use two production-level AV systems, Apollo 5.5 and Autoware 1.12.0. as baselines.
Both AV systems are heavily tested to drive physical vehicles in the real world~\cite{apollodebut,10.1109/ICCPS.2018.00035}.
COLA doesn't modify the specific algorithm implementation in the system, as it may hurt the correctness and accuracy of the systems.
Instead, COLA includes about 1.5k LoC changes, mostly in the scheduling systems~\footnote{We will open-source COLA upon publication of this paper.}.

\begin{figure*}
    \centering

    \begin{subfigure}[b]{.19\textwidth}
        \includegraphics[width=\linewidth]{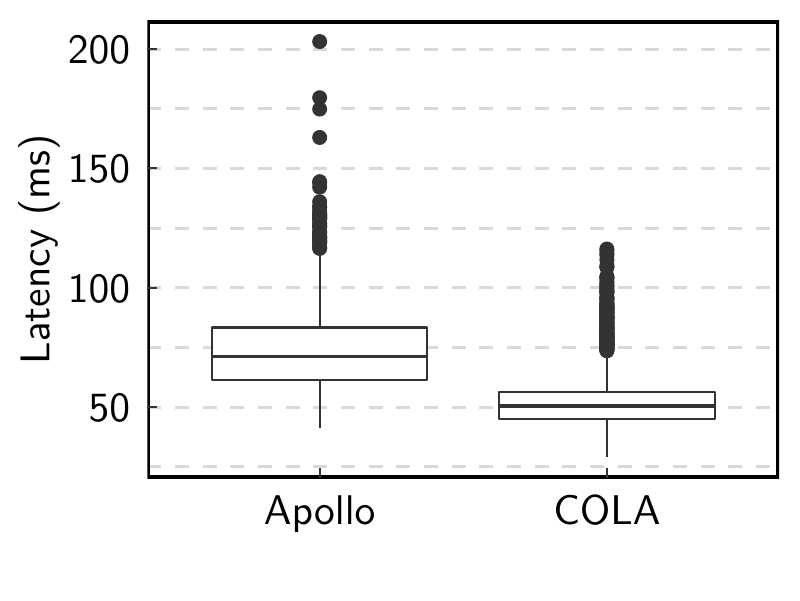}
        \caption{The Performance improvements of COLA in Apollo.}
        \label{fig:all_perf}
    \end{subfigure}
    \hfill
    \begin{subfigure}[b]{.19\textwidth}
        \includegraphics[width=\linewidth]{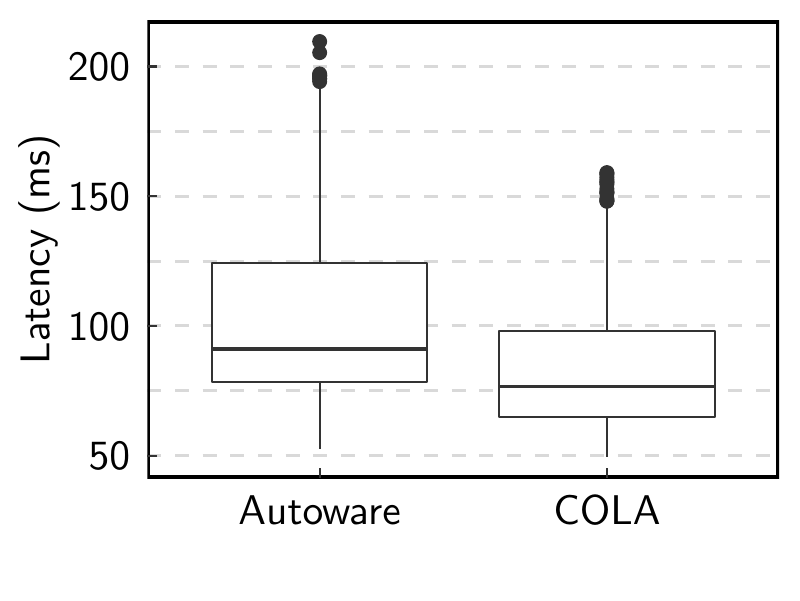}
        \caption{COLA performance improvements in Autoware.}
        \label{fig:all_perf_autoware}
    \end{subfigure}
    \hfill
    \begin{subfigure}[b]{.19\textwidth}
        \includegraphics[width=\linewidth]{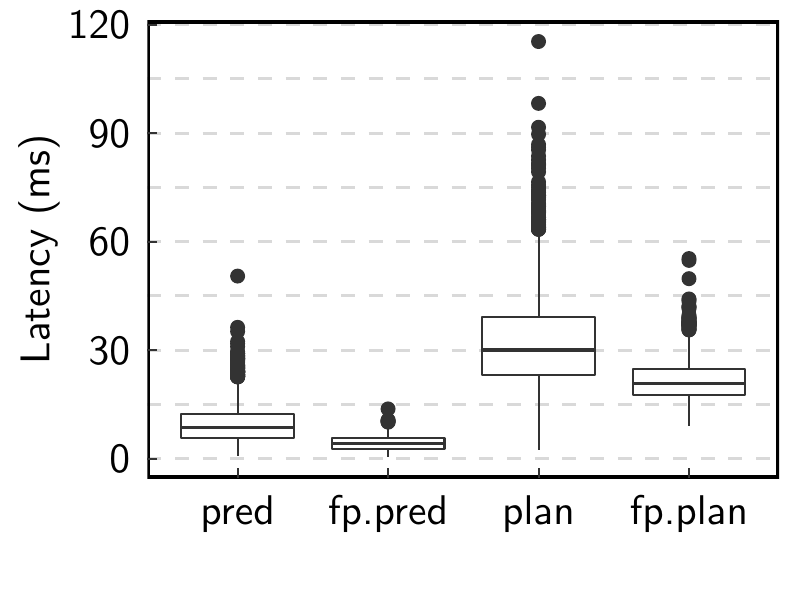}
        \caption{The latency distribution between fastpath and original.}
        \label{fig:prune_comparison}
    \end{subfigure}
    \hfill
    \begin{subfigure}[b]{.19\textwidth}
        \includegraphics[width=\linewidth]{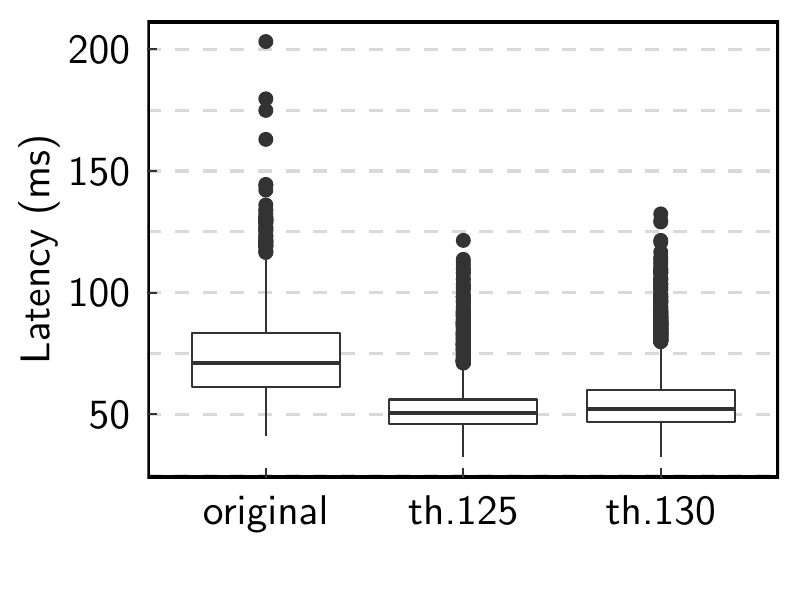}
        \caption{The end-to-end latency with varying latency budgets.}
        \label{fig:all_comparison}
    \end{subfigure}
    \hfill
    \begin{subfigure}[b]{.19\textwidth}
        \includegraphics[width=\linewidth]{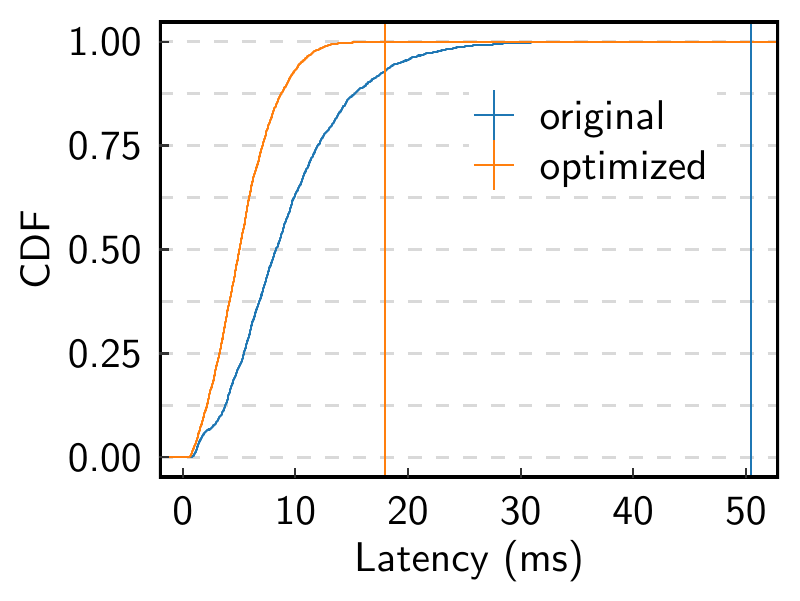}
        \caption{Tail latency before/after proactive processing.  }
        \label{fig:optimized_cdf}
    \end{subfigure}
    \\
    \begin{subfigure}[t]{.19\textwidth}
        \includegraphics[width=\linewidth]{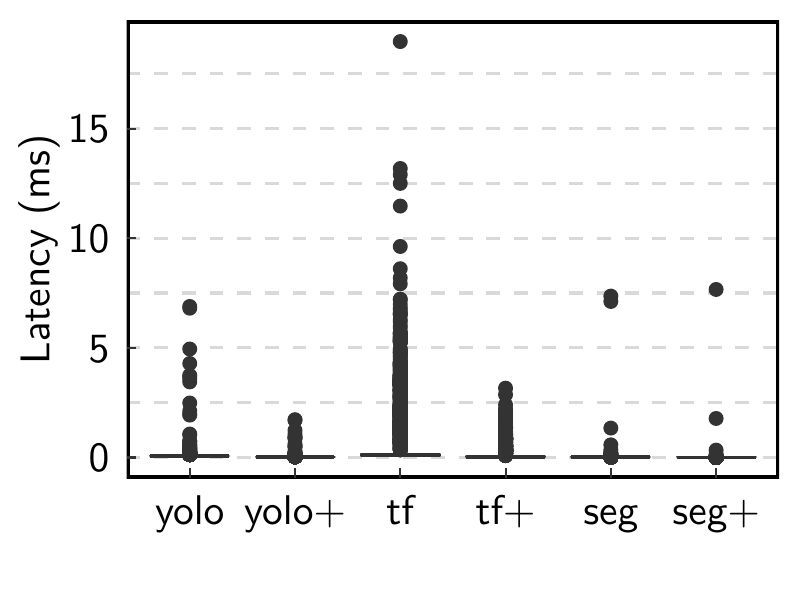}
        \caption{Latency distribution before/after optimization.}
        \label{fig:dy_memcpy}
    \end{subfigure}
    \hfill
    \begin{subfigure}[t]{.19\textwidth}
        \includegraphics[width=\linewidth]{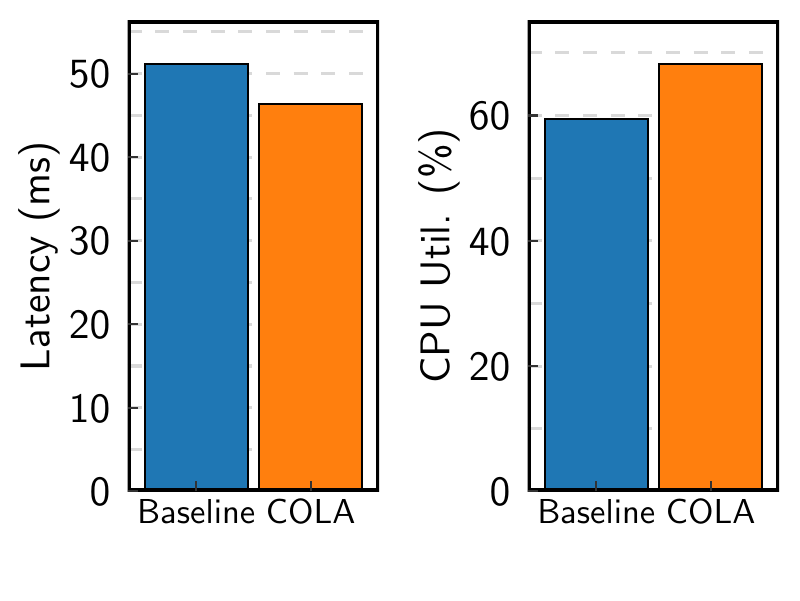}
        \caption{Performance and utilization improvements by best-effort stealing.}
        \label{fig:steal}
    \end{subfigure}
    \hfill
    \begin{subtable}[t]{.19\textwidth}
        \vspace{-5em}
        \resizebox{.95\textwidth}{!}{
            \scriptsize \bfseries \ttfamily
            \begin{tabular}{|l|c|}
                \hline
                System        & Collision \\ \hline
                Apollo        & 0         \\ \hline
                Apollo-COLA   & 0         \\ \hline
                Autoware      & 21        \\ \hline
                Autoware-COLA & 16        \\ \hline
            \end{tabular}
        }
        \captionsetup{skip=20pt}
        \caption{Collision cases of different systems with/without COLA.}
        \label{tab:safety_normal}
    \end{subtable}
    \hfill
    \begin{subfigure}[t]{.19\textwidth}
        \includegraphics[width=\linewidth]{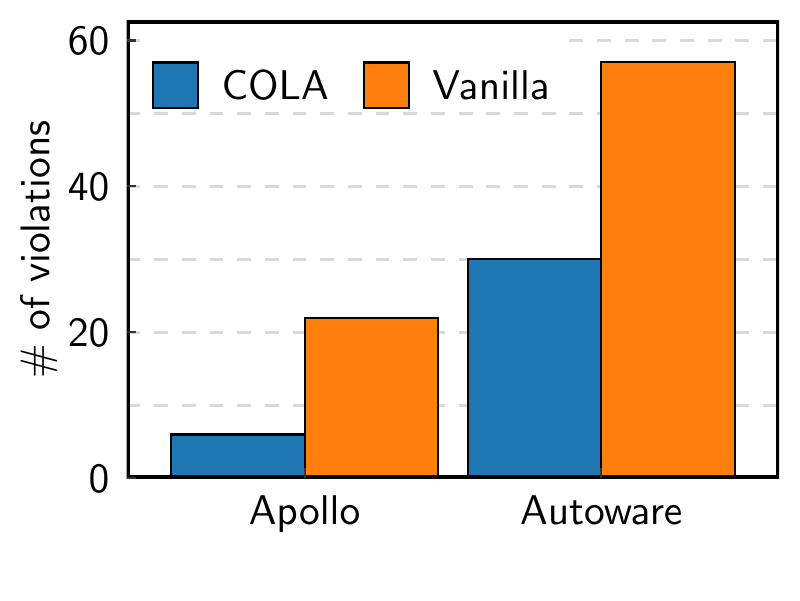}
        \caption{Safety violation times in tested scenarios.}
        \label{fig:violation}
    \end{subfigure}
    \hfill
    \begin{subfigure}[t]{.19\textwidth}
        \includegraphics[width=\linewidth]{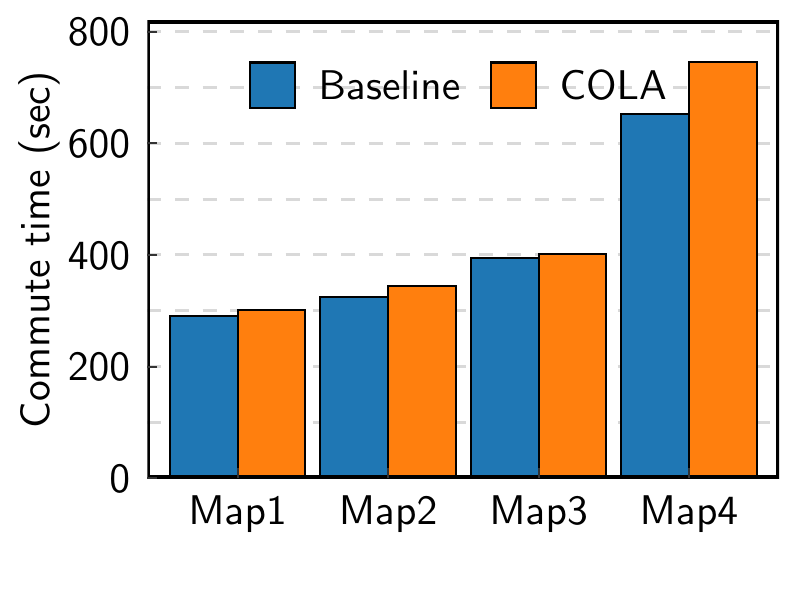}
        \caption{The driving time between baseline and COLA-Apollo.}
        \label{fig:city}
    \end{subfigure}
    \vspace{-10pt}
    \caption{COLA evaluation results.}
    \label{fig:cola-eval}
    \vspace{-10pt}
\end{figure*}

\begin{figure}

    \begin{subfigure}[b]{.23\textwidth}
        \includegraphics[width=\linewidth]{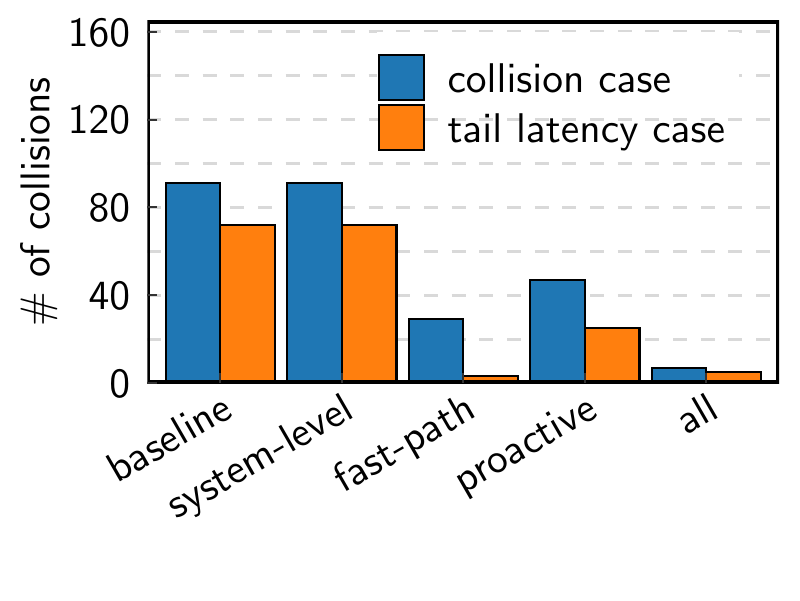}
        \caption{Crash cases with vanilla Apollo and COLA.}
        \label{fig:safety_eval}
    \end{subfigure}
    \hfill
    \begin{subfigure}[b]{.23\textwidth}
        \includegraphics[width=\linewidth]{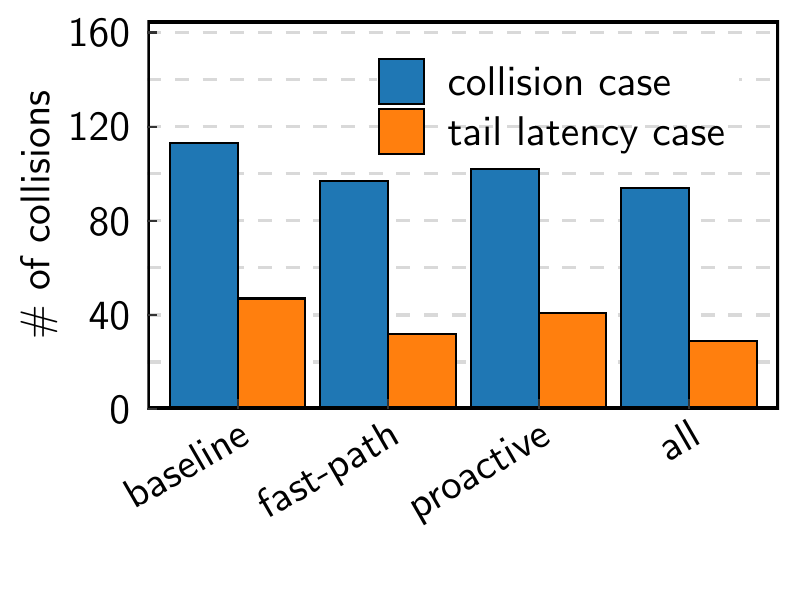}
        \caption{Crash cases with vanilla Autoware and COLA.}
        \label{fig:safety_eval_autoware}
    \end{subfigure}
    \vspace{-10pt}
    \caption{Safety analysis and ablation study for COLA on Apollo and Autoware}
    \label{fig:safety_eval_both}
    \vspace{-20pt}
\end{figure}

\subsection{Performance Evaluation}

In this section, we evaluate the performance improvement of our design. To avoid the impacts of the variation brought by traffic, we are using the same scenario to test the performance with/without our design.
We implement COLA designs on Apollo and Autoware and show the overall system performance in Figure~\ref{fig:all_perf} and
Figure~\ref{fig:all_perf_autoware}. In our tested scenarios, COLA-Apollo can achieve a mean latency of 52.38 ms and a worst-case latency of 123.93ms, compared with original latency of 73.73 ms and worst-case latency of 203.16ms; COLA-Autoware can achieve a mean latency of 82.77 ms and a worst-case latency of 159.1ms, compared with original latency of 101.73 ms and worst-case latency of 209.61 ms.

\para{Fast path.}
Figure \ref{fig:prune_comparison} shows the latency distribution between the fast path and normal path.
We can see that the fast path achieves lower and more predictable latency.
We also observe that the latency of succeeding planning modules also falls dramatically after selecting the fast path, because both modules have lower latency when processing fewer obstacles.

The fast path determines to take the fast path or not based on the specified object-level deadlines.
Figure \ref{fig:all_comparison} summarizes the result with different deadlines. \texttt{th-125} indicates we use a deadline of 125 ms and so forth. We achieve a worst-case end-to-end latency of 121.4 ms, below 125 ms. when the threshold is higher, the tail latency is lower. 

\para{Proactive processing.} Figure \ref{fig:optimized_cdf} shows the performance of proactive processing. We find that our asynchronous design can bound the worst-case latency within 20 ms. It also reduces the mean latency by 1.7$\times$.
Figure \ref{fig:dy_memcpy} shows improved GPU data transfer performance. \texttt{yolo}, \texttt{tf}, \texttt{seg} shows the original GPU data transferring of YOLO object detector, traffic light detection, and LiDAR segmentation. the \texttt{yolo+} shows the performance on our ahead-of-time data transferring optimization and so forth. The result shows that
our design bounds memory transfering of traffic light and yolo task under 7 ms.

\para{Best-effort work stealing.} Figure \ref{fig:steal} shows 7.8\% performance gains in LiDAR segmentation modules and 15\% improvement in CPU utilization. 


\subsection{Safety Study}
We evaluate our design in 8329 driving scenarios~\footnote{We describe the scenario details in Appendix~\ref{appendix:scenario}}, including two categories: (1) normal traffic scenarios, to see if our design brings unexpected side effects.
(2) corner case scenarios, to show if our designs can improve safety with faster reactions.

\para{Normal Traffic.}  
First, we collect 8000 driving scenarios (25 seconds each) with normal traffic in Carla Traffic Simulation~\cite{carlatraffic}.
We compare the collision rate of Apollo, COLA-Apollo (optimized Apollo), Autoware and COLA-autoware (optimized Autoware) in Table~\ref{tab:safety_normal}.
The result shows that COLA designs do not add extra crash cases.

\para{Safety Violation Analysis}
Based on our analysis in Section \ref{section:latencyrequirements}, AVs can drive too close to other traffic agents (including vehicles and pedestrians) caused by tail latency in computing systems.
Such safety violations may not lead to collisions but are still dangerous.
In our simulation, we compute the safe longitudinal and lateral distance based on the RSS model~\cite{intelrss}, and collect the times when AV fails to keep safe distances from other vehicles. 
Figure \ref{fig:violation} shows that COLA significantly reduces the violation times, reducing 72.7\% and 47.3\% accidents in Apollo and Autoware, respectively.

\para{Corner-case Scenario Evaluation.}
We also explore the safety of our design in corner-case scenarios. Those scenarios require faster reaction time and therefore are more challenging for AV systems.
COLA curated 329 corner-case scenarios.
We also run the original Apollo/Autoware and the modified version with 329 corner-case scenarios and collect the collision cases.
The results are shown in Figure~\ref{fig:safety_eval_both}. The \textit{all} result shows those techniques collectively reduce about 93\% crash cases in Apollo and 61\% in Autoware, illustrating the effectiveness of our approach.

\subsection{Ablation Study}
In Figure \ref{fig:safety_eval_both}, we perform an ablation study to understand how each design in COLA affects safety improvement.
\texttt{system-level}, \texttt{fast-path} and \texttt{proactive} refers to applying each design individually. \texttt{all} refers to the combination of all designs.
The simulation results show that none of our designs bring extra safety concerns. 
One important reason is that our design is not intrusive, we are simply reusing the current AV framework and building \textit{proactive} and \textit{fast-path} based on the existing functionality.

\subsection{Accuracy Degradation Analysis}

Our design fastpath may lead to sub-optimal driving decisions (e.g. more conservative), which potentially hurts the driving efficiency.
To quantify such an effect, we run experiments to measure the commuting time of the unmodified Apollo and Apollo with COLA designs. The AVs are instructed to drive from the same source to the same destination with the same traffic in the simulator, each repeating 10 times.
Figure \ref{fig:city} shows the average commuting time in different maps, which are predefined in CARLA simulator with various road conditions. The traffic becomes heavier from \texttt{Map1} to \texttt{Map4}.
We can see that COLA has an average 6.43\% driving efficiency penalty, especially in \texttt{Map4} with heavier traffic.
\section{Related Work}

\para{Real-time systems.}
Real-time systems are designed to serve incoming requests and complete them within a predictable latency bound~\cite{10.1145/844128.844144}, it typically models real-time tasks with worst-case task latency and recurring intervals and finds an optimal scheduling algorithm based on the workload characteristics~\cite{Carpenter2004ACO}. 
Our work serves as a complement to the previous real-time system research, in favor of an algorithm-system
codesign approach to mitigate the tail latency problem in industrial AV systems.


\para{Tail Latency.}
A large body of previous research work aims at mitigating tail latency or making it predictable in different layers of modern computing systems~\cite{256983, 10.1145/2670979.2670988, 254354, 10.1145/3132747.3132774,10.1145/844128.844144, 10.5555/1855741.1855747,40801}. Level-4 AVs pose new challenges due to dynamic latency requirements, resources constraint, and sensitivity to traffic~\cite{10.1145/3296957.3173191, 9251251}. 
Pylot is a research prototype AV system investigating the latency-accuracy tradeoff~\cite{Gog2021PylotAM, 10.1145/3492321.3519576}. However, it is questionable if the system is applicable in real-world settings.

\para{AV Performance Optimization.}
A number of works study the AV system performance in terms of bottleneck analysis~\cite{9251251}, latency-accuracy tradeoff~\cite{Gog2021PylotAM}, design constraints such as power in AV frameworks~\cite{10.1145/2155620.2155650, 254354}, and also time constraints~\cite{Luo2019TimeCA,10.1145/3492321.3519576}.
\section{Conclusion}

We propose COLA, a Level-4 AV system characterization and optimization framework.
We characterize the tail latency in Level-4 AV systems in terms of latency source and safety implication.
We make seven observations in three categories, including AV reaction time issues, tail latency characterizations, and system-level implications on AV safety and stability.
Based on these observations, we propose a set of AV system designs, including adaptive dataflow, proactive processing, and best-effort work stealing, to mitigate tail latency.
The evaluations demonstrate that our designs significantly reduce the tail latency effects in current state-of-the-art AV systems, thereby improving driving safety.


\clearpage

\bibliographystyle{plain}
\bibliography{content/ref.bib}

\begin{thebibliography}{10}

\bibitem{cyberrt}
Apollo cyber rt faqs.
\newblock
  \url{https://github.com/ApolloAuto/apollo/blob/master/docs/cyber/CyberRT_FAQs.md}.

\bibitem{apollortkernel}
Apollo realtime kernel patch.
\newblock \url{https://github.com/ApolloAuto/apollo-kernel}.

\bibitem{apolloscenario}
Apollo simulation: A comprehensive solution for the development of autonomous
  vehicles.
\newblock \url{https://developer.apollo.auto/platform/simulation.html}.

\bibitem{nhtsasafety}
Automated driving systems: A vision for safety.
\newblock
  \url{https://www.nhtsa.gov/sites/nhtsa.gov/files/documents/13069a-ads2.0_090617_v9a_tag.pdf}.

\bibitem{apollo}
Baidu apollo.
\newblock \url{https://github.com/ApolloAuto/apollo}.

\bibitem{apollodebut}
Baidu apollo debuts the first level-4 autonomous buses in china.
\newblock
  \url{https://www.futurecar.com/4158/Baidu-Apollo-Debuts-the-First-Level-4-Autonomous-Buses-in-China/}.

\bibitem{carlaleaderboard}
Carla autonomous driving leaderboard.
\newblock \url{https://leaderboard.carla.org/}.

\bibitem{crashsurvey}
National motor vehicle crash causation survey.
\newblock
  \url{https://crashstats.nhtsa.dot.gov/Api/Public/ViewPublication/811059}.

\bibitem{nhtsascenario}
Pre-crash scenario typology for crash avoidance research.
\newblock
  \url{file:///home/haolan/Downloads/Pre-Crash_Scenario_Typology-Final_PDF_Version_5-2-07.pdf}.

\bibitem{TaxonomyAD}
Taxonomy and definitions for terms related to driving automation systems for
  on-road motor vehicles.

\bibitem{teslaaccident3}
Tesla autopilot system found probably at fault in 2018 crash.
\newblock
  \url{https://www.nytimes.com/2020/02/25/business/tesla-autopilot-ntsb.html}.

\bibitem{congestionsurvey}
Traffic congestion: The problem and how to deal with it.
\newblock
  \url{https://repositorio.cepal.org/bitstream/handle/11362/37898/1/LCG2199P_en.pdf}.

\bibitem{carlatraffic}
Traffic simulation in carla.
\newblock
  \url{https://carla.readthedocs.io/en/latest/ts_traffic_simulation_overview}.

\bibitem{teslaaccident2}
Two killed in tesla crash with no driver at the wheel.
\newblock
  \url{https://www.forbes.com/sites/jonathanponciano/2021/04/18/driverless-tesla-behind-crash-that-killed-two-in-texas-officials-believe/?sh=2a2cfc044824
  }.

\bibitem{UberL4}
Uber advanced technologies group: A principled approach to safety.
\newblock \url{https://docs.huihoo.com/car/Uber-ATGSafety-Report-2018.pdf}.

\bibitem{uberaccident}
Uber's self-driving operator charged over fatal crash.
\newblock \url{https://www.bbc.com/news/technology-54175359}.

\bibitem{waymo}
Waymo:enabling autonomous.
\newblock
  \url{https://www.mobileye.com/our-technology/mobileye-enabling-autonomous/}.

\bibitem{teslaaccident}
‘it happened so fast’: Inside a fatal tesla autopilot accident.
\newblock
  \url{https://www.nytimes.com/2021/08/17/business/tesla-autopilot-accident.html}.

\bibitem{256983}
Cake: Enabling high-level slos on shared storage systems.
\newblock Boston, MA, June 2012. {USENIX} Association.

\bibitem{199317}
Mart{\'\i}n Abadi, Paul Barham, Jianmin Chen, Zhifeng Chen, Andy Davis, Jeffrey
  Dean, Matthieu Devin, Sanjay Ghemawat, Geoffrey Irving, Michael Isard,
  Manjunath Kudlur, Josh Levenberg, Rajat Monga, Sherry Moore, Derek~G. Murray,
  Benoit Steiner, Paul Tucker, Vijay Vasudevan, Pete Warden, Martin Wicke, Yuan
  Yu, and Xiaoqiang Zheng.
\newblock {TensorFlow}: A system for {Large-Scale} machine learning.
\newblock In {\em 12th USENIX Symposium on Operating Systems Design and
  Implementation (OSDI 16)}, pages 265--283, Savannah, GA, November 2016.
  USENIX Association.

\bibitem{Bateni2019PredictableDR}
Soroush Bateni and Cong Liu.
\newblock Predictable data-driven resource management: an implementation using
  autoware on autonomous platforms.
\newblock {\em 2019 IEEE Real-Time Systems Symposium (RTSS)}, pages 339--352,
  2019.

\bibitem{254354}
Soroush Bateni and Cong Liu.
\newblock Neuos: A latency-predictable multi-dimensional optimization framework
  for dnn-driven autonomous systems.
\newblock In {\em 2020 {USENIX} Annual Technical Conference ({USENIX} {ATC}
  20)}, pages 371--385. {USENIX} Association, July 2020.

\bibitem{9251251}
Pedro H.~E. Becker, José~María Arnau, and Antonio González.
\newblock Demystifying power and performance bottlenecks in autonomous driving
  systems.
\newblock In {\em 2020 IEEE International Symposium on Workload
  Characterization (IISWC)}, pages 205--215, 2020.

\bibitem{Carbone2015ApacheFS}
Paris Carbone, Asterios Katsifodimos, Stephan Ewen, Volker Markl, Seif Haridi,
  and Kostas Tzoumas.
\newblock Apache flink{\texttrademark}: Stream and batch processing in a single
  engine.
\newblock {\em IEEE Data Eng. Bull.}, 38:28--38, 2015.

\bibitem{Carpenter2004ACO}
John Carpenter, Shelby~H. Funk, Philip Holman, A.~Srinivasan, James~H.
  Anderson, and Sanjoy Baruah.
\newblock A categorization of real-time multiprocessor scheduling problems and
  algorithms.
\newblock In {\em Handbook of Scheduling}, 2004.

\bibitem{mmcv}
MMCV Contributors.
\newblock {MMCV: OpenMMLab} computer vision foundation.
\newblock \url{https://github.com/open-mmlab/mmcv}, 2018.

\bibitem{40801}
Jeffrey Dean and Luiz~André Barroso.
\newblock The tail at scale.
\newblock {\em Communications of the ACM}, 56:74--80, 2013.

\bibitem{DBLP:journals/corr/abs-2202-02215}
Wenhao Ding, Chejian Xu, Mansur Arief, Haohong Lin, Bo~Li, and Ding Zhao.
\newblock A survey on safety-critical driving scenario generation - {A}
  methodological perspective.
\newblock {\em CoRR}, abs/2202.02215, 2022.

\bibitem{DBLP:journals/corr/abs-1808-05819}
Nemanja Djuric, Vladan Radosavljevic, Henggang Cui, Thi Nguyen, Fang{-}Chieh
  Chou, Tsung{-}Han Lin, and Jeff Schneider.
\newblock Motion prediction of traffic actors for autonomous driving using deep
  convolutional networks.
\newblock {\em CoRR}, abs/1808.05819, 2018.

\bibitem{Dosovitskiy17}
Alexey Dosovitskiy, German Ros, Felipe Codevilla, Antonio Lopez, and Vladlen
  Koltun.
\newblock {CARLA}: {An} open urban driving simulator.
\newblock In {\em Proceedings of the 1st Annual Conference on Robot Learning},
  pages 1--16, 2017.

\bibitem{aaareaction}
Paweł Droździel, Sławomir Tarkowski, Iwona Rybicka, and Rafał Wrona.
\newblock Drivers ’reaction time research in the conditions in the real
  traffic.
\newblock {\em Open Engineering}, 10(1):35--47, 2020.

\bibitem{fan2018baidu}
Haoyang Fan, Fan Zhu, Changchun Liu, Liangliang Zhang, Li~Zhuang, Dong Li,
  Weicheng Zhu, Jiangtao Hu, Hongye Li, and Qi~Kong.
\newblock Baidu apollo em motion planner, 2018.

\bibitem{10.1145/3302424.3303977}
Alireza Farshin, Amir Roozbeh, Gerald~Q. Maguire, and Dejan Kosti\'{c}.
\newblock Make the most out of last level cache in intel processors.
\newblock In {\em Proceedings of the Fourteenth EuroSys Conference 2019},
  EuroSys '19, New York, NY, USA, 2019. Association for Computing Machinery.

\bibitem{Fremont2020FormalST}
Daniel~J. Fremont, Edward Kim, Yash Pant, S.~Seshia, Atul Acharya, Xantha
  Bruso, Paul Wells, Steve Lemke, Q.~Lu, and Shalin Mehta.
\newblock Formal scenario-based testing of autonomous vehicles: From simulation
  to the real world.
\newblock {\em 2020 IEEE 23rd International Conference on Intelligent
  Transportation Systems (ITSC)}, pages 1--8, 2020.

\bibitem{gan2020eudoxus}
Yiming Gan, Yu~Bo, Boyuan Tian, Leimeng Xu, Wei Hu, Shaoshan Liu, Qiang Liu,
  Yanjun Zhang, Jie Tang, and Yuhao Zhu.
\newblock Eudoxus: Characterizing and accelerating localization in autonomous
  machines, 2020.

\bibitem{intelrss}
Bernd Gassmann, Fabian Oboril, Cornelius Buerkle, Shuang Liu, Shoumeng Yan,
  Maria~Soledad Elli, Ignacio Alvarez, Naveen Aerrabotu, Suhel Jaber, Peter van
  Beek, Darshan Iyer, and Jack Weast.
\newblock Towards standardization of av safety: C++ library for responsibility
  sensitive safety.
\newblock In {\em 2019 IEEE Intelligent Vehicles Symposium (IV)}, 2019.

\bibitem{10.1145/844128.844144}
Ashvin Goel, Luca Abeni, Charles Krasic, Jim Snow, and Jonathan Walpole.
\newblock Supporting time-sensitive applications on a commodity os.
\newblock {\em SIGOPS Oper. Syst. Rev.}, 36(SI):165–180, December 2003.

\bibitem{10.1145/3492321.3519576}
Ionel Gog, Sukrit Kalra, Peter Schafhalter, Joseph~E. Gonzalez, and Ion Stoica.
\newblock D3: A dynamic deadline-driven approach for building autonomous
  vehicles.
\newblock In {\em Proceedings of the Seventeenth European Conference on
  Computer Systems}, EuroSys '22, page 453–471, New York, NY, USA, 2022.
  Association for Computing Machinery.

\bibitem{Gog2021PylotAM}
Ionel Gog, Sukrit Kalra, Peter Schafhalter, Matthew~A. Wright, Joseph Gonzalez,
  and Ion Stoica.
\newblock Pylot: A modular platform for exploring latency-accuracy tradeoffs in
  autonomous vehicles.
\newblock {\em 2021 IEEE International Conference on Robotics and Automation
  (ICRA)}, pages 8806--8813, 2021.

\bibitem{10.1145/3132747.3132774}
Mingzhe Hao, Huaicheng Li, Michael~Hao Tong, Chrisma Pakha, Riza~O. Suminto,
  Cesar~A. Stuardo, Andrew~A. Chien, and Haryadi~S. Gunawi.
\newblock Mittos: Supporting millisecond tail tolerance with fast rejecting
  slo-aware os interface.
\newblock In {\em Proceedings of the 26th Symposium on Operating Systems
  Principles}, SOSP '17, page 168–183, New York, NY, USA, 2017. Association
  for Computing Machinery.

\bibitem{DBLP:journals/corr/abs-2107-08142}
Ashesh Jain, Luca~Del Pero, Hugo Grimmett, and Peter Ondruska.
\newblock Autonomy 2.0: Why is self-driving always 5 years away?
\newblock {\em CoRR}, abs/2107.08142, 2021.

\bibitem{10.1109/ICCPS.2018.00035}
Shinpei Kato, Shota Tokunaga, Yuya Maruyama, Seiya Maeda, Manato Hirabayashi,
  Yuki Kitsukawa, Abraham Monrroy, Tomohito Ando, Yusuke Fujii, and Takuya
  Azumi.
\newblock Autoware on board: Enabling autonomous vehicles with embedded
  systems.
\newblock In {\em Proceedings of the 9th ACM/IEEE International Conference on
  Cyber-Physical Systems}, ICCPS '18, page 287–296. IEEE Press, 2018.

\bibitem{avhardware}
Tobias Kessler, Julian Bernhard, Martin Buechel, Klemens Esterle, Patrick Hart,
  Daniel Malovetz, Michael Le, Frederik Diehl, Thomas Brunner, and Alois Knoll.
\newblock Bridging the gap between open source software and vehicle hardware
  for autonomous driving.
\newblock pages 1612--1619, 06 2019.

\bibitem{10.1145/2670979.2670988}
Jialin Li, Naveen~Kr. Sharma, Dan R.~K. Ports, and Steven~D. Gribble.
\newblock Tales of the tail: Hardware, os, and application-level sources of
  tail latency.
\newblock In {\em Proceedings of the ACM Symposium on Cloud Computing}, SOCC
  '14, page 1–14, New York, NY, USA, 2014. Association for Computing
  Machinery.

\bibitem{10.1145/3296957.3173191}
Shih-Chieh Lin, Yunqi Zhang, Chang-Hong Hsu, Matt Skach, Md~E. Haque, Lingjia
  Tang, and Jason Mars.
\newblock The architectural implications of autonomous driving: Constraints and
  acceleration.
\newblock {\em SIGPLAN Not.}, 53(2):751–766, March 2018.

\bibitem{icl:16}
Kevin London, Shirley Moore, Phil Mucci, Keith Seymour, and Richard Luczak.
\newblock The papi cross-platform interface to hardware performance counters.
\newblock In {\em Department of Defense Users{\textquoteright} Group Conference
  Proceedings}, Biloxi, Mississippi, 2001-06 2001.

\bibitem{Luo2019TimeCA}
Yujia Luo.
\newblock Time constraints and fault tolerance in autonomous driving systems.
\newblock 2019.

\bibitem{10.1145/2155620.2155650}
Jason Mars, Lingjia Tang, Robert Hundt, Kevin Skadron, and Mary~Lou Soffa.
\newblock Bubble-up: Increasing utilization in modern warehouse scale computers
  via sensible co-locations.
\newblock In {\em Proceedings of the 44th Annual IEEE/ACM International
  Symposium on Microarchitecture}, MICRO-44, page 248–259, New York, NY, USA,
  2011. Association for Computing Machinery.

\bibitem{Quigley2009ROSAO}
M.~Quigley.
\newblock Ros: an open-source robot operating system.
\newblock In {\em ICRA 2009}, 2009.

\bibitem{DBLP:journals/corr/abs-1808-10703}
Atsushi Sakai, Daniel Ingram, Joseph Dinius, Karan Chawla, Antonin Raffin, and
  Alexis Paques.
\newblock Pythonrobotics: a python code collection of robotics algorithms.
\newblock {\em CoRR}, abs/1808.10703, 2018.

\bibitem{ShalevShwartz2017OnAF}
Shai Shalev-Shwartz, Shaked Shammah, and Amnon Shashua.
\newblock On a formal model of safe and scalable self-driving cars.
\newblock {\em ArXiv}, abs/1708.06374, 2017.

\bibitem{DBLP:journals/corr/abs-1711-05805}
Guowei Wan, Xiaolong Yang, Renlan Cai, Hao Li, Hao Wang, and Shiyu Song.
\newblock Robust and precise vehicle localization based on multi-sensor fusion
  in diverse city scenes.
\newblock {\em CoRR}, abs/1711.05805, 2017.

\bibitem{6856582}
Junqing Wei, Jarrod~M. Snider, Tianyu Gu, John~M. Dolan, and Bakhtiar Litkouhi.
\newblock A behavioral planning framework for autonomous driving.
\newblock In {\em 2014 IEEE Intelligent Vehicles Symposium Proceedings}, pages
  458--464, 2014.

\bibitem{frenettrajectory}
Moritz Werling, Julius Ziegler, Sören Kammel, and Sebastian Thrun.
\newblock Optimal trajectory generation for dynamic street scenarios in a
  frenet frame.
\newblock pages 987 -- 993, 06 2010.

\bibitem{frenetoptimal}
Moritz Werling, Julius Ziegler, Sören Kammel, and Sebastian Thrun.
\newblock Optimal trajectory generation for dynamic street scenarios in a
  frenet frame.
\newblock pages 987 -- 993, 06 2010.

\bibitem{wu2019detectron2}
Yuxin Wu, Alexander Kirillov, Francisco Massa, Wan-Yen Lo, and Ross Girshick.
\newblock Detectron2.
\newblock \url{https://github.com/facebookresearch/detectron2}, 2019.

\bibitem{xu2020data}
Kecheng Xu, Xiangquan Xiao, Jinghao Miao, and Qi~Luo.
\newblock Data driven prediction architecture for autonomous driving and its
  application on apollo platform, 2020.

\bibitem{10.5555/1855741.1855747}
Ting Yang, Tongping Liu, Emery~D. Berger, Scott~F. Kaplan, and J.~Eliot~B.
  Moss.
\newblock Redline: First class support for interactivity in commodity operating
  systems.
\newblock In {\em Proceedings of the 8th USENIX Conference on Operating Systems
  Design and Implementation}, OSDI'08, page 73–86, USA, 2008. USENIX
  Association.

\bibitem{9251973}
B.~{Yu}, W.~{Hu}, L.~{Xu}, J.~{Tang}, S.~{Liu}, and Y.~{Zhu}.
\newblock Building the computing system for autonomous micromobility vehicles:
  Design constraints and architectural optimizations.
\newblock In {\em 2020 53rd Annual IEEE/ACM International Symposium on
  Microarchitecture (MICRO)}, pages 1067--1081, 2020.

\bibitem{180560}
Matei Zaharia, Mosharaf Chowdhury, Tathagata Das, Ankur Dave, Justin Ma, Murphy
  McCauly, Michael~J. Franklin, Scott Shenker, and Ion Stoica.
\newblock Resilient distributed datasets: A {Fault-Tolerant} abstraction for
  {In-Memory} cluster computing.
\newblock In {\em 9th USENIX Symposium on Networked Systems Design and
  Implementation (NSDI 12)}, pages 15--28, San Jose, CA, April 2012. USENIX
  Association.

\bibitem{zhao2019towards}
Hengyu Zhao, Yubo Zhang, Pingfan Meng, Hui Shi, Li~Erran Li, Tiancheng Lou, and
  Jishen Zhao.
\newblock Towards safety-aware computing system design in autonomous vehicles.
\newblock {\em arXiv preprint arXiv:1905.08453}, 2019.

\bibitem{9304602}
Hengyu Zhao, Yubo Zhang, Pingfan Meng, Hui Shi, Li~Erran Li, Tiancheng Lou, and
  Jishen Zhao.
\newblock Safety score: A quantitative approach to guiding safety-aware
  autonomous vehicle computing system design.
\newblock In {\em 2020 IEEE Intelligent Vehicles Symposium (IV)}, pages
  1479--1485, 2020.

\bibitem{DBLP:journals/corr/abs-2103-01882}
Jinyun Zhou, Rui Wang, Xu~Liu, Yifei Jiang, Shu Jiang, Jiaming Tao, Jinghao
  Miao, and Shiyu Song.
\newblock Exploring imitation learning for autonomous driving with feedback
  synthesizer and differentiable rasterization.
\newblock {\em CoRR}, abs/2103.01882, 2021.

\end{thebibliography}

\clearpage
\appendix

\section{Tested Scenario.}
\label{appendix:scenario}

A large line of prior work explores AV safety testing and validation~\cite{nhtsascenario, ShalevShwartz2017OnAF}. Among them, scenario-based testing is a widely accepted method, by testing AV with diverse driving scenarios~\cite{DBLP:journals/corr/abs-2202-02215, DBLP:journals/corr/abs-2103-01882, frenetoptimal}.
The tested scenario sets need to be: (1) diverse enough to include possible situations in the real world. and (2) safety-critical to cause a collision if handled improperly.

To this end, we curated a scenario dataset of 8329 scenarios, mainly from three sources: (1) CARLA leaderboard, a widely used safety-evaluation benchmark for AVs~\cite{carlaleaderboard}. (2) NHTSA precrash reports~\cite{nhtsasafety}. The National Highway Traffic Safety Administration (NHTSA) reports pre-crash scenarios in 44 categories, depicting vehicle movements and dynamics as well as the critical event immediately prior to a crash, which are often used in vehicle safety research.
(3) Apollo Scenario dataset: Baidu Apollo curates many common scenarios for the safety and correctness testing ~\cite{apolloscenario}.
The following list shows our testing scenario and its brief description:

\begin{itemize}
\item \textbf{Vehicle Following}: the AV is following another vehicle in the same lane, trying to keep a safe distance.
\item \textbf{Lane Changing}: the AV is to change to another lane with vehicles, or another vehicle in the neighbor lane is trying to cut in, AV should yield to avoid crashes.
\item \textbf{Negotiation at the intersection \& roundabout}: including protected and unprotected turn, left turn yielding on green.  
\item \textbf{Merging Lane}: going into the merging lane.
\item \textbf{Cyclists \& pedestrians}: how to handle cyclists and pedestrians.

\item \textbf{Traffic Rules}: testing if the AV is able to properly handle traffic signs and traffic lights.
\item \textbf{Nudging}: the AV is to nudge to pass a badly parked vehicle.
\item \textbf{Emergency}: testing if the AV is able to properly pullover to the curbside, when necessary.
\item \textbf{Road Curve}: the AV should avoid the road edge. 
\item \textbf{Occlusion}: an unexpected pedestrian or AV shows up behind occlusion. 
\item \textbf{Reckless driver}: the other vehicle not following traffic rules: run the red light, make an illegal U-turn.
\end{itemize}

For the robustness of our testing, we also mutate (by random sampling or adding perturbation) other simulation parameters such as weather, vehicle initial positions, velocities and poses.

\section{Simulation Configuration.}
\label{appendix:simulation}

We use Carla, a high-fidelity simulator in our experiments.
Previous work compares the on-road testing and the simulator and shows that 62\% of unsafe behaviors in the simulation lead to real crashes and 93\% of safe behaviors in the simulation are also safe in real life~\cite{Fremont2020FormalST}. The comparison experiment shows that our simulation results are able to transfer to road tests well.

\begin{figure}
    \centering
    \includegraphics[width=0.45\textwidth]{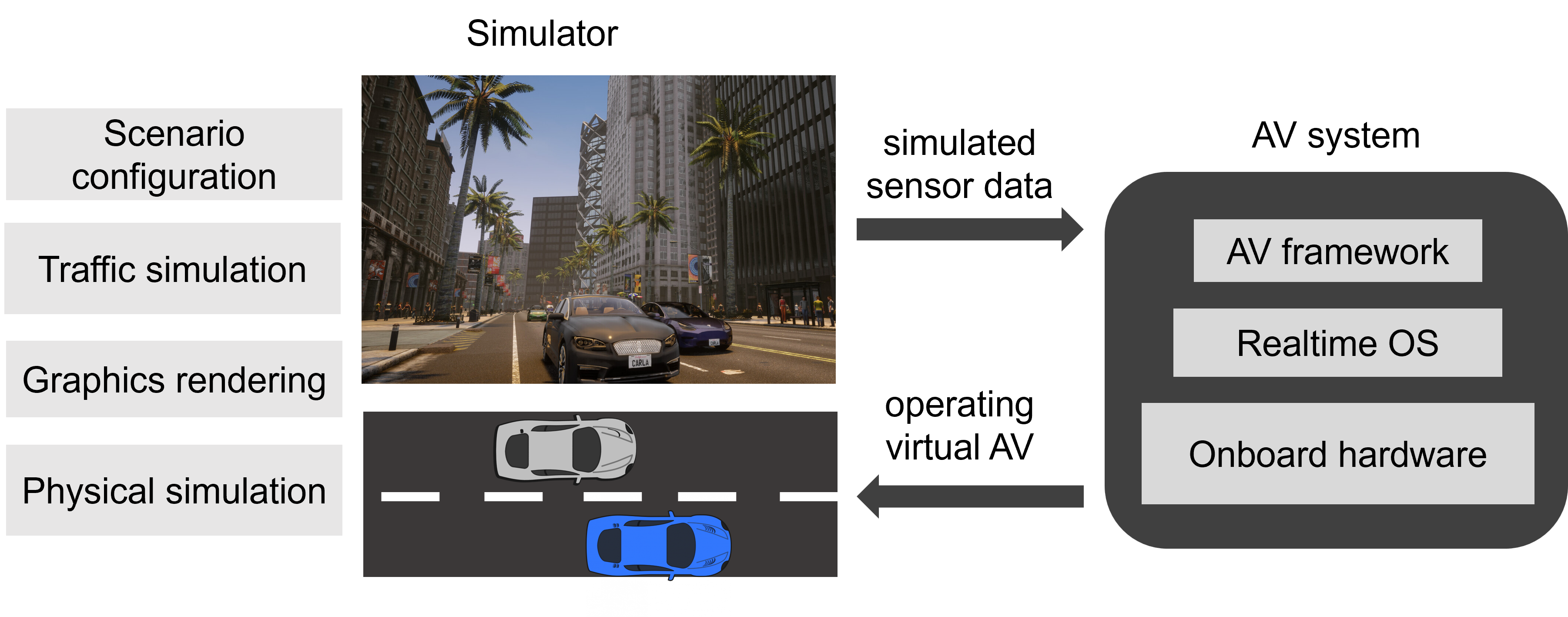}

    \caption{Experimental setup of COLA: The AV system is running on real hardware to operate the virtual vehicle.}

    \label{fig:simulator}
\end{figure}

Our experiment setup is in Figure \ref{fig:simulator},
the simulator will simulate a virtual traffic scenario, including the configuration of the virtual world (the road, traffic, weather, etc) and the events (e.g., a vehicle from the neighbor lane ready to overtake) based on the physical model. The simulator will also simulate high fidelity raw sensor data to the AV stack using graphics techniques (e.g., rendering). We run real industrial AV software and hardware along with the simulator.
The AV software takes the sensor input into a modular computation pipeline and controls the virtual AV.
We also use the same sensor configuration recommended by Apollo~\cite{apolloscenario}. The specific sensor configuration is listed in the Table~\ref{table:sensor}.

\begin{table}[]
\centering
\footnotesize
\begin{tabular}{|c|c|c|c|}
\hline
Category &
  Name &
  Number &
  Description \\ \hline
\multirow{3}{*}{Camera} &
  \begin{tabular}[c]{@{}c@{}}front \\ camera\end{tabular} &
  1 &
  \begin{tabular}[c]{@{}c@{}}6 mm \\ camera\end{tabular} \\ \cline{2-4} 
 &
  \begin{tabular}[c]{@{}c@{}}side \\ camera\end{tabular} &
  2 &
  \begin{tabular}[c]{@{}c@{}}6 mm \\ camera\end{tabular} \\ \cline{2-4} 
 &
  \begin{tabular}[c]{@{}c@{}}front \\ main camera\end{tabular} &
  1 &
  \begin{tabular}[c]{@{}c@{}}12 mm \\ camera\end{tabular} \\ \hline
\multirow{2}{*}{LiDAR} &
  \begin{tabular}[c]{@{}c@{}}top \\ LiDAR\end{tabular} &
  1 &
  \begin{tabular}[c]{@{}c@{}}Velodyne 128 \\ laser beams, \\ on top of the vehicle\end{tabular} \\ \cline{2-4} 
 &
  \begin{tabular}[c]{@{}c@{}}front \\ LiDAR\end{tabular} &
  1 &
  \begin{tabular}[c]{@{}c@{}}Velodyne 16 laser beams, \\ in the front\end{tabular} \\ \hline
Radar &
  \begin{tabular}[c]{@{}c@{}}the front/rear \\ Radar\end{tabular} &
  2 &
  N/A \\ \hline
\end{tabular}
\caption{The sensor configuration of our tested vehicle.}
\label{table:sensor}
\end{table}

\end{document}